\documentclass[journal=acs,manuscript=article]{achemso}
\setkeys{acs}{journal=acs,keywords = true,etalmode=truncate,maxauthors=0} 
\usepackage{chemformula} 
\usepackage[T1]{fontenc} 

\usepackage{microtype}
\usepackage{graphicx}
\usepackage{subcaption}
\usepackage{amsmath,amsfonts,amssymb,amsthm}
\usepackage{mathtools}
\usepackage{commath}
\usepackage{textcomp} 
\usepackage{booktabs} 
\usepackage{float} 
\usepackage{amssymb} 
\usepackage{bm} 
\usepackage{hyperref}
\usepackage{sidecap}
\usepackage{longtable}
\usepackage{algorithm}
\usepackage{algorithmic}
\usepackage{fancyhdr}
\usepackage[final]{changes} 
\definechangesauthor[name=AES, color=green]{AES}

\newcommand{\beginsupplement}{%
        \setcounter{table}{0}
        \renewcommand{\thetable}{S-\arabic{table}}%
        \setcounter{figure}{0}
        \renewcommand{\thefigure}{S-\arabic{figure}}
        \setcounter{section}{0}
        \renewcommand{\thesection}{S-\arabic{section}}%
        \setcounter{page}{1}
        \renewcommand{\thepage}{S-\arabic{page}}
     }

\mciteErrorOnUnknownfalse
\author{Alexander E. Siemenn}
\affiliation{Department of Mechanical Engineering, Massachusetts Institute of Technology, Cambridge, MA 02139, USA}
\email{asiemenn@mit.edu}
\author{Evyatar Shaulsky}
\affiliation{Department of Chemical Engineering, Northeastern University, Boston, MA 02115, USA}
\author{Matthew Beveridge}
\affiliation{Department of Electrical Engineering and Computer Science, Massachusetts Institute of Technology, Cambridge, MA 02139, USA}
\author{Tonio Buonassisi}
\affiliation{Department of Mechanical Engineering, Massachusetts Institute of Technology, Cambridge, MA 02139, USA}
\author{Sara M. Hashmi}
\affiliation{Department of Chemical Engineering, Northeastern University, Boston, MA 02115, USA}
\email{s.hashmi@northeastern.edu}
\author{Iddo Drori}
\affiliation{Department of Electrical Engineering and Computer Science, Massachusetts Institute of Technology, Cambridge, MA 02139, USA}
\email{idrori@mit.edu}

\title{A Machine Learning and Computer Vision Approach to \added{Rapidly} Optimize Multiscale Droplet Generation}


\begin{document}


\begin{abstract}

Generating droplets from a continuous stream of fluid requires precise tuning of a device to find optimized control parameter conditions. \added{It is analytically intractable to compute the necessary control parameter values of a droplet-generating device that produces optimized droplets. Furthermore, as the length scale of the fluid flow changes, the formation physics and optimized conditions that induce flow decomposition into droplets also change. Hence, a single proportional integral derivative controller is too inflexible to optimize devices of different length scales or different control parameters, while classification machine learning techniques take days to train and require millions of droplet images. Therefore, the question is posed, can a single method be created that universally optimizes multiple length-scale droplets using only a few data points and is faster than previous approaches? In this paper, a Bayesian optimization and computer vision feedback loop is designed to quickly and reliably discover the control parameter values that generate optimized droplets within different length-scale devices. This method is demonstrated to converge on optimum parameter values using 60 images in only 2.3 hours, $30\times$ faster than previous approaches. Model implementation is demonstrated for two different length-scale devices: a milliscale inkjet device and a microfluidics device.}
\deleted{current physics-based droplet optimization tools are sensitive to the scale of the droplet being optimized and also require \textit{a priori} knowledge of the system physics to perform optimization. We propose a Bayesian machine learning tool integrated with computer vision to optimize droplets within various devices at multiple length scales while requiring no prior domain knowledge of the system. This control method is validated on two devices of different length scales and different droplet formation physics: an inkjet device at the milliscale and a mircofluidics device at the microscale.}
\end{abstract}


\section{Introduction}

\begin{figure}
\centering
\includegraphics[width=1\columnwidth]{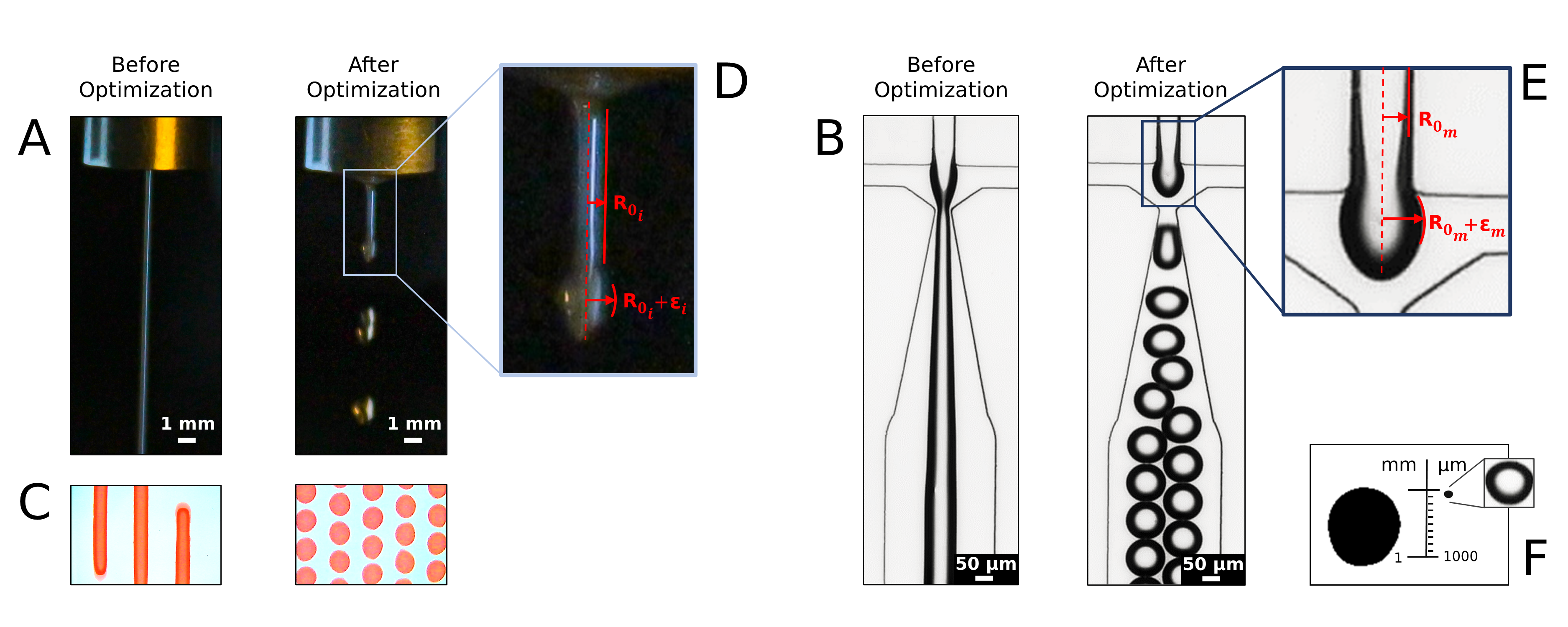}
\caption{Fluid flow before and after control parameter optimization to form droplets. (A) Rayleigh instability and (B) capillary instability forming after optimization for the inkjet and microfluidics devices, respectively. (C) Deposited droplet patterns from the inkjet device. The images show a case from before optimization, using Latin hypercube sampled control parameter values, and a case from after optimization. Our model learns how to optimize droplet, which indirectly optimizes (D) Rayleigh and (E) capillary instabilities within the fluid stream. Rayleigh instability forms within the inkjet device by applying piezoelectric-driven perturbations, $\varepsilon_i$, to a water flow of radius $R_{0_i}$ being acted on by gravity in ambient air. Capillary instability forms within the microfluidic device by applying viscous stress-driven perturbations, $\varepsilon_m$, from the oil medium to a water flow of radius $R_{0_m}$. (F) Length scale comparison of an inkjet-generated droplet (left in mm) and a microfluidic-generated droplet (right in \textrm{$\mu$}m).}
\label{fig:rayleigh}
\end{figure}


Generating discrete and uniform droplets from a fluid stream requires the fine-tuning of experimental conditions to transition the flow from a continuous, stable jet to a separated, unstable stream of droplets \cite{Song2020}. Rayleigh-Plateau instability governs the separation of a fluid stream into discrete droplets through gravity-driven perturbations and capillary instability governs the separation of a fluid stream into discrete droplets through pressure-driven perturbations \cite{Papageorgiou1995}. \added{Analytical relationships between governing parameters and droplet characteristics simply do not exist due to the complex, non-linear physical relationships between all forces acting on the fluid at various length scales \cite{gu2011droplets}. Thus, it is analytically intractable to determine the parameter values that produce optimized droplets, hence, requiring experimentation. Typically, researchers tune experimental controls until they reach the desired droplet characteristics  \cite{Mei2004, Song2020, anna2006microscale, Derby2011}.  Our method provides an efficient way to find the appropriate values of experimental control parameters that achieve the desired outcome without developing a device-specific control model. Thus, we are no longer restricted to using our method only on devices with similar governing physics. We demonstrate that the method successfully optimizes droplets generated by devices that have both dissimilar control parameters and droplet length-scales, with few modifications. Furthermore, the novelty of our approach lies not only in its flexible application to devices with different governing physics and control parameters, but also in it requiring no upfront time to train, enabling researchers to quickly apply our method to their data and get results within hours rather than days \cite{Chu2019}.

Prior works develop proportional integral derivative (PID) controllers or classification machine learning to predict or control droplet states \cite{miller2010microfluidic, Chu2019}. However, both of these approaches are inflexible as they either require specific models to simulate the physics of a specific device, in the case of PID, or require training on large datasets of data specific to a single device, in the case of classification. Additionally, it has been shown that an estimated three days of upfront training is necessary to effectively classify and predict the droplet states of a microfluidic device \cite{Chu2019}. Our method requires no upfront training, instead, our algorithm learns the dynamics between input control parameters iteratively as the researcher performs experiments. We demonstrate that after initialization, our method requires only four rounds of experimentation with a batch size of 10 before converging on an optimum. Resulting in an average total optimization time of 2 hours, compared to previously reported $>70$ hours via supervised classification \cite{Chu2019}.

Flexible machine learning (ML) processes, such as Bayesian optimization (BO), have the promise to greatly improve the efficiency of exploring parameter spaces by saving expensive lab time and resources as they are generalizable to parameter spaces of varying dimensionality and require no upfront model training since they use a Gaussian Process (GP) backbone \cite{Drugowitsch2019, Barnes2011,miller2010microfluidic,Hennig2012, Seeger2004}. In this paper, we develop a universal BO and computer vision feedback loop to quickly and reliably discover Rayleigh and capillary unstable conditions in different droplet-generating devices with different governing control parameters and physics \cite{ohnesorge2019formation,baroud2010dynamics}. We demonstrate the functionality of our method on two devices: (1) an milliscale inkjet droplet-generator with the controllable parameters of (a) pressure, (b) actuation frequency, and (c) translation speed and (2) a microfluidic droplet-generator with the controllable parameters of (a) water pressure and (b) oil pressure, shown in Fig. \ref{fig:rayleigh} and Fig. \ref{fig:inkjet-diagram}.

}

\deleted{Generating discrete and uniform droplets from a fluid stream requires the fine-tuning of experimental conditions to transition the flow from a continuous, stable jet to a separated, unstable stream of droplets \cite{Song2020}. Rayleigh-Plateau instability governs the separation of a fluid stream into discrete droplets through gravity-driven perturbations and capillary instability governs the separation of a fluid stream into discrete droplets through pressure-driven perturbations \cite{Papageorgiou1995}. These fluid stream instabilities are controllable by varying the experimental conditions of the fluid devices and by achieving the right combination of conditions, the stream is decomposed into discrete droplets \cite{Mei2004, Song2020}. However, the exact combination of conditions that produce discrete droplets is often analytically expensive to compute due to the complex, non-linear physical relationships between all forces acting on the fluid at various length scales \cite{gu2011droplets}. Moreover, the discovery of droplet-generating conditions via experimentation often requires trial-and-error experimentation, which does not always achieve the most optimal settings. Therefore, in this study, we demonstrate the use of a Bayesian machine learning (ML) and computer vision feedback loop to reliably discover the values of a device's control parameters that form Rayleigh/capillary instabilities to generate an optimized droplet flow \cite{Drugowitsch2019,Hennig2012, Seeger2004}. The novelty of this model is that it is scale-invariant and requires no \textit{a prioi} domain knowledge of the jetting system or the physics to optimize droplets at various length scales.

We utilize this Bayesian ML model to discover Rayleigh unstable conditions in an inkjet device that generates droplets at the milliscale and then use same tool to discovery capillary unstable conditions in a microfluidic device that generates droplets at the microscale, shown in Fig. \ref{fig:rayleigh} \cite{ohnesorge2019formation,baroud2010dynamics}. While the dependence of drop size and number on a single flow rate in a defined device geometry can be predicted to a certain extent by modeling, the larger parameter space of device conditions may be cumbersome to explore manually in the laboratory \cite{anna2006microscale, Derby2011}. Additionally, the parameter space of each droplet device has a different number of dimensions and different formation physics, for example, in this study, the milliscale device has a three-dimensional parameter space: \textit{fluid pressure $\times$ piezoelectric actuation frequency $\times$ nozzle translation speed} while the microscale device has a two-dimensional parameter space: \textit{water pressure $\times$ oil pressure}.

ML processes, such as Bayesian inference, have the promise to greatly improve the efficiency of exploring parameter space and save expensive lab time resources as they are generalizable to parameter spaces of varying dimensionality \cite{Drugowitsch2019, Barnes2011,miller2010microfluidic}. Closed loop PID systems have been used to form droplets of a desired size, but these methods require \textit{a priori} knowledge of how droplet size depends on experimental control parameters \cite{miller2010microfluidic}.  As such, closed loop PID systems are best suited to explore systems in which the governing physics is straightforward and the control parameter space is limited. Droplet formation, however, exhibits non-linear dependence on a variety of factors, and it is often desired to explore a wide range of experimental parameters \cite{ohnesorge2019formation, baroud2010dynamics}. We provide a method that offers the ability to control droplet size without requiring \textit{a priori} knowledge of droplet size variation as a function of the control parameters.
}

The challenge faced in optimizing droplets at different length scales is in designing a tool robust enough to predict the physics of each device parameter space to make meaningful and reliable predictions of optimized printing conditions. The challenge is addressed in this paper through our physical model-agnostic approach of optimization, which implements computer vision to detect the droplet flows and converts them into numerical representations of the parameter space for the Bayesian inference algorithm to learn. The significance of developing a tool that rapidly optimizes droplets at multiple length scales is in providing researchers with a universal tool for tuning devices that requires little customization or ML expertise. Generating these optimized droplets has importance across several application fields depending on the length scale of the droplet.

\subsection{Prior Work on Millimeter-scale Droplets}

Depositing millimeter-scale droplets onto a substrate is a process useful for high-throughput characterization of material, most notably for finding optimized semiconductor materials that maximize efficiency or stability  \cite{Bash2020, Langner2020}. Semiconductors such as perovskites have vast and complex composition spaces which makes it challenging to discover optimum compositions \cite{Sun2019, SUN2021, Ren2020}. Thus, using discrete inkjet-deposited droplets of varying semiconductor compositions for high-throughput experimental characterization elicits an accelerated search of this vast composition space for an optimum composition. However, studies such as \textit{Bash et al.,} (2020) rely on a domain expert to fine-tune experimental conditions that establish flow instability control prior to semiconductor characterization and, hence, require and understanding of system physics prior to optimization. Rather, in this study, the non-linear physical relationships between control parameters are iteratively learned by the probabilistic ML surrogate model as data is collected.

\subsection{Prior Work on Micrometer-scale Droplets}

Creating microspheres with a high surface area to volume ratio is beneficial for a large variety of applications such as materials generation, bio(chemical) analysis, polymeric microcapsules, and more \cite{fridman2021high,ding2019recent, shang2017emerging}. Industries like food science, biomedical, and cosmetics use microcapsules as materials delivery vehicles with the ability to tune the capsule wall chemistry for slow or triggered release of the capsulated material \cite{kaufman2015soft,bah2020fabrication, huang2017generation}. For each of these unique applications, a different microdroplet volume is required. However, these droplet volumes and structures are a function of several physical parameters like the interfacial tension between the immiscible fluids, their viscosities, the microfluidic device geometry, and fluid flow rates \cite{sesen2017droplet, joanicot2005droplet}. \replaced{

Existing literature uses supervised neural networks to inform device geometry or to classify different control states for droplet generation. This method requires a large quantity of training data and requires an estimated 70 hours for model training.  However, a gap still exists on how ML can be applied to optimize the device control parameters using GP techniques that do not require \textit{a priori} training and can, in turn, operate much faster \cite{lashkaripour2021machine, Chu2019}.}{Existing literature uses ML to inform device geometry for droplet generation, however, a gap still exists on how ML can be applied to optimize the device control parameters \cite{lashkaripour2021machine}.}. Thus, in this study, we use Bayesian inference methods to optimize the microspheres using the control parameters of water and oil pressures directly, rather than developing a PID model or supervised neural network to simulate the physical parameters and their non-linear relationships.

\subsection{Main Findings}

We demonstrate accurate and repeatable discovery of Rayleigh and capillary unstable regions to generate droplets of high yield and high circularity within two fluid devices at different length scales using Bayesian optimization and computer vision. For each of these fluid devices, three trials of Bayesian optimization were run using three different decision policy acquisition functions in which all trials converged to learning similar bounds for generating high circularity and high yield droplet structures. 
\replaced{Convergence on optimized control parameters is attained for both devices using the same method within 2 hours while synthesizing 60 samples only.}{The developed Bayesian optimization and computer vision process loop attains consistent convergence on these learned parameter space topologies using only 60 experimental samples, including the initialization set.}

\section{Methods}

\begin{figure}[ht]
\centering
\includegraphics[width=\textwidth]{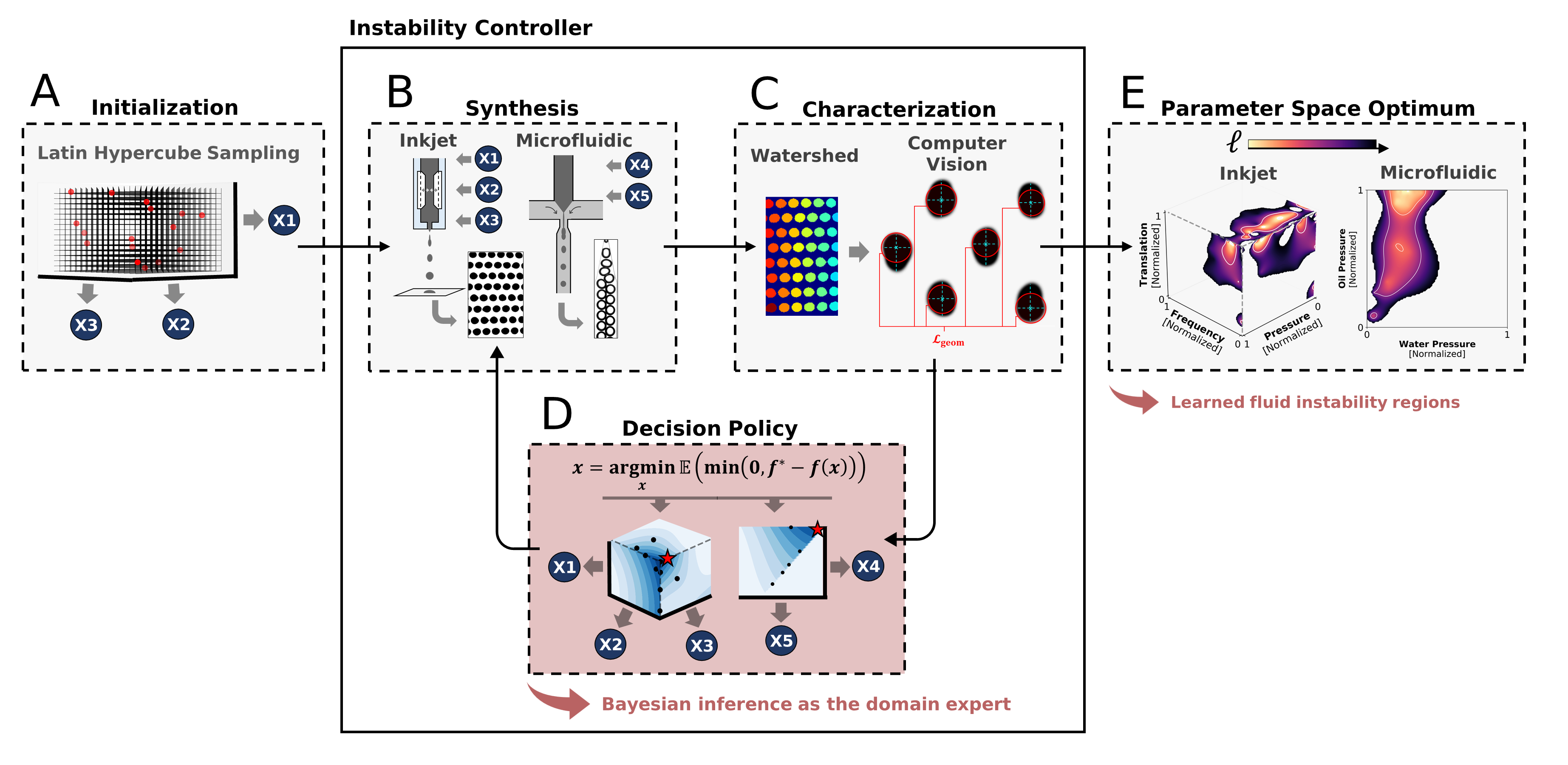}
\caption{\replaced{Control feedback loop of the hardware-algorithm interface used for droplet optimization.}{The process flow of the computer vision BO tool.} (A) The BO experiment is initialized using LHS, which samples control parameter values from the device parameter space. (B) For the first loop, LHS parameter values are used to print and image droplets. For proceeding rounds, learned parameter values by BO are used to print and image droplets. (C) Computer vision is used to score each droplet image and assign a label that quantifies droplet circularity and yield. \added{The hardware (B) and computer vision (C) subsystems interface with each other by simply providing an image of the output droplets to the computer vision code.} (D) The label and parameter values from the current loop and all prior loops are input into the BO decision policy \added{directly from the computer vision algorithm}, which informs the next round of parameter values to print. The iterative retraining of BO from processes (B)--(D) autonomously drives the control of the Rayleigh and capillary instability phenomena without \textit{a priori} domain knowledge of a physical model. (E) Convergence on optimized control parameter values is assessed after four loops of iteratively retraining BO with new experimentally generated data.}
\label{fig:workflow}
\end{figure}

\added{Fig. \ref{fig:workflow} illustrates the workflow presented in this paper for multiscale droplet optimization. The user collects a set of initialization images of droplets generated from the device over a range of experimental control parameters (\textit{e.g.}, fluid pressures, nozzle frequency). The optimization software analyzes the data and generates a new set of control parameters to be used in the device, and the process is then repeated.  Within only a few iterations, the efficient algorithm converges to a set of control parameters to produce the desired drops.}

\subsection{Device Hardware}

\begin{figure}[p]
\centering
\begin{subfigure}{0.6\textwidth}
\added[id=AES, comment={Replace inkjet lab photo with CAD drawing for clarity and labeling purposes.}]{\includegraphics[width=\textwidth]{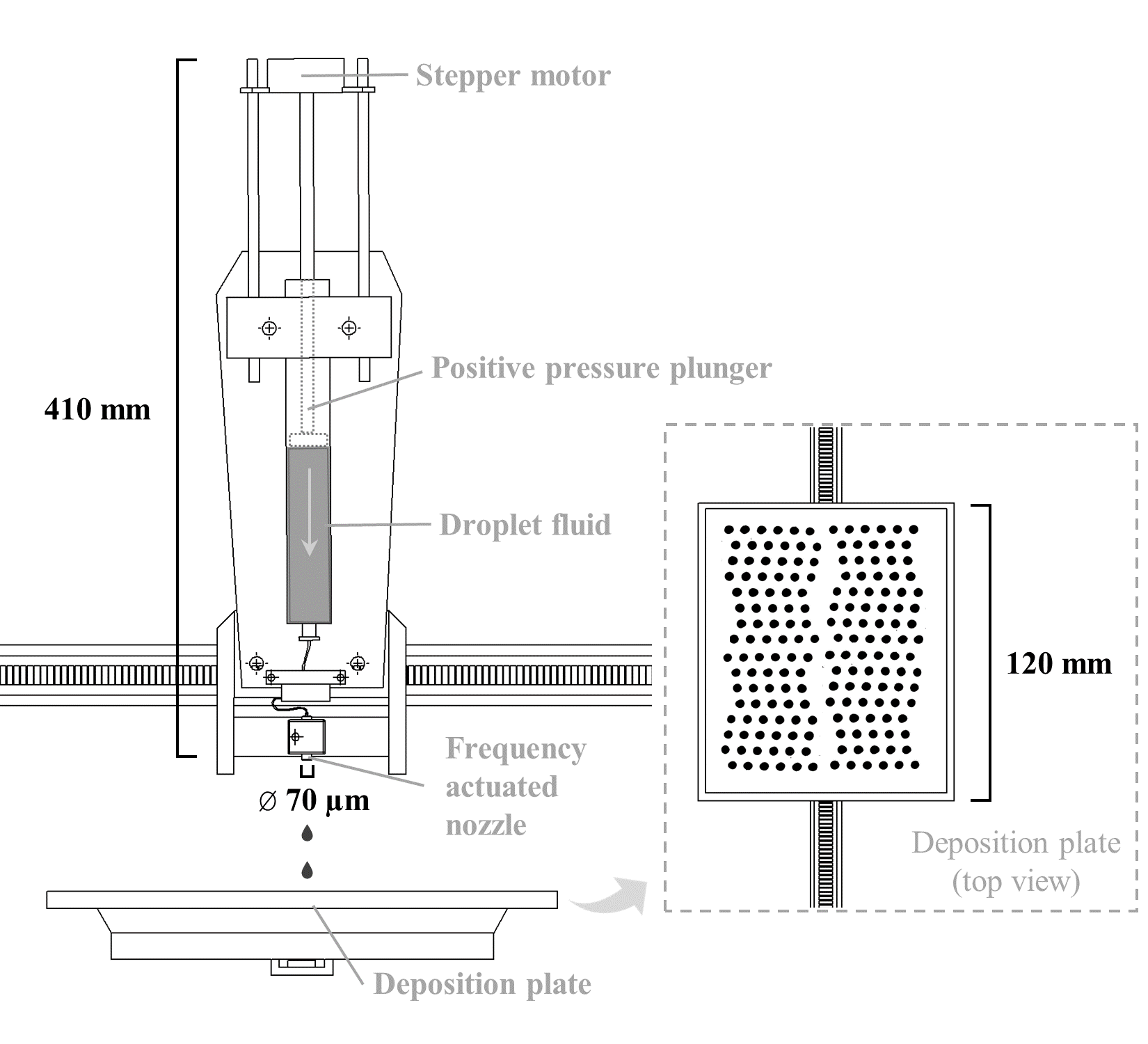}}
\caption{\added{Inkjet Device Architecture}}
\label{fig:inkjet_hardware}
\end{subfigure}
\begin{subfigure}{0.55\textwidth}
\added[id=AES, comment={Replace microfluidics lab photo with CAD drawing for clarity and labeling purposes.}]{\includegraphics[width=\textwidth]{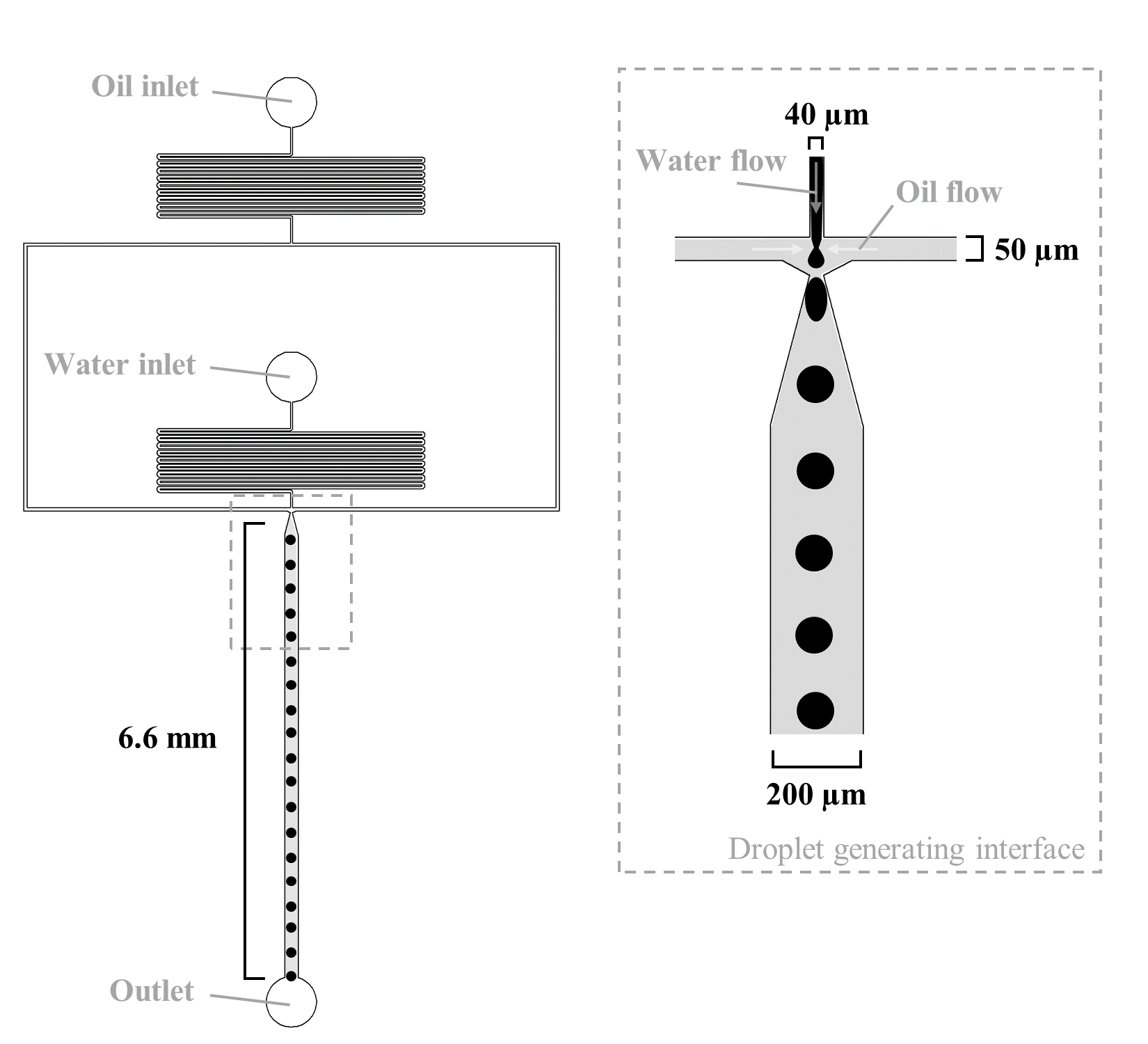}}
\caption{\added{Microfluidics Chip Architecture}}
\label{fig:micro_hardware}
\end{subfigure}
\caption{\added{Droplet-generating device hardware used in this study. (a) Inkjet device with inset image showing the top view of the deposition plate and the (b) microfluidics device with inset image showing the droplet generating interface. For the inkjet device, the positive pressure motor controls the fluid pressure in the vials; the frequency of the actuated nozzle controls the rate of droplet deposition; the speed of the laterally translating plate controls the droplet deposition placement. For the microfluidics device, the channel is flowed with mineral oil from the oil line as the suspending fluid; oil pinches off water from the water line to form droplets; all fluid exits through the outlet. The entire microfluidics device has a height of 30 \textrm{$\mu$m}.}}
\label{fig:inkjet-diagram}
\end{figure}

\subsubsection{Inkjet}

\added{The inkjet device has three possible control parameters: (1) fluid pressure, (2) piezoelectric actuation frequency, and (3) nozzle translation speed, of which we know the discrete values of at all times, but do not know the combination of these values that produce optimized droplets. These three parameters influence the shape and yield of droplets deposited on a plate \cite{Derby2011}. Pressurizing the fluid within the pipes drives both the radius and the flow rate of the stream for a fixed orifice size; our inkjet device operates at pressures limits of 0.03--0.15 MPa. Applying an alternating current electrical signal to the piezoelectric material within the pipe actuates a membrane at a given frequency to induce perturbations, $\varepsilon$, within the fluid stream \cite{ohnesorge2019formation, baroud2010dynamics}; our inkjet device operates at frequency limits of 1.0--600.0 Hz using. The fluid is ejected from a nozzle which translates in 1D above a deposition site, which also translates in 1D perpendicular to the nozzle, driving 2D deposition of the fluid onto the deposition plate; our inkjet device operates at translation speed limits of 10--360 mm/s. None of these three control parameters are artificially constrained, the limits are defined by the physical limits of the hardware only. We keep both the fluid (dyed water) and nozzle diameter (70 \textrm{$\mu$}m) constant for all experiments, shown in Fig. \ref{fig:inkjet_hardware}. The intrinsic fluid properties are constant and we operate at flow velocities between 0.8--4 m/s, given the range of operating pressure. For every printed sample, these known values of the control parameters are appended to the image and normalized to a range of $[0, 1]$ before being fed into the BO algorithm.

The printable fluid parameter space is governed by the fluid density, surface tension, viscosity, as well as the driving pressure and nozzle geometry \cite{McKinley2011, CLANET1999, Tai2008}. While analytical relationships between these variables and the output droplet circularity and yield are intractable, the droplet properties can be described by dimensionless relationships between forces: (1) Ohnesorge number (Oh), which compares viscous forces to inertia and surface tension, (2) Weber number (We), which compares inertial forces to surface tension, and (3) Reynolds number (Re), which compares inertial to viscous forces \cite{McKinley2011, CLANET1999}.}
\begin{equation}
    \textrm{Oh}=\frac{\mu}{\sqrt{\rho \sigma d}}, \qquad
    \textrm{We}=\frac{\rho v^2 a}{\sigma}, \qquad
    \textrm{Re}=\frac{\rho v d}{\mu},
\end{equation}
\replaced{where $\mu$ is the fluid dynamic viscosity, $\rho$ is the fluid density, $\sigma$ is the surface tension, $d$ is the jet diameter, $a$ is the droplet diameter, and $v$ is the characteristic flow velocity. $\textrm{Oh}$ depends on fluid properties and nozzle diameter only, and so is fixed in our system at $\textrm{Oh}\approx 0.01$. This value falls in the range typically considered to facilitate satellite-drop formation; we observe satellite drops in approximately 10\% of the inkjet samples \cite{McKinley2011}. Moreover, we determine the ranges of the variable dimensionless numbers to be $\textrm{We} \approx$ 1--200 and $\textrm{Re} \approx$ 10--50, which are further discussed in the results. Gravity is negligible in our system because the Bond number $=\frac{\rho g d^2}{\sigma}\approx 0.001$, where $g$ is gravitational acceleration, implying that capillary forces dominate the gravitational forces. Interestingly, despite operating in the regime of satellite drops, the data-driven BO approach efficiently explores and optimizes the parameter space for desired drop properties.
}{Our model represents the inkjet device as a function of three control parameters: (1) fluid pressure, (2) piezoelectric actuation frequency, and (3) nozzle translation speed, which we know the discrete values of at all times. These three parameters influence the shape and yield of droplets deposited on a plate \cite{Derby2011}. Pressurizing the fluid within the pipes drives both the radius and the flow rate of the stream for a fixed orifice size; our inkjet device operates at pressures limits of 0.03--0.15 MPa. Applying an alternating current electrical signal to the piezoelectric material within the pipe actuates a membrane at a given frequency to induce perturbations, $\varepsilon$, within the fluid stream \cite{ohnesorge2019formation, baroud2010dynamics}; our inkjet device operates at frequency limits of 1.0--600.0 Hz using. The fluid is ejected from a nozzle which translates above a deposition site, driving the point of deposition of the fluid; our inkjet device operates at translation speed limits of 10--360 mm/s. We use the same fluid and small nozzle orifice for all experiments, the fluid being dyed water, and a nozzle orifice of 70 \textrm{$\mu$}m. The intrinsic fluid properties are constant and we operate at very high flow velocity for all operating pressures. For every printed sample, these known values of the control parameters are appended to the image and normalized to a range of $[0, 1]$ before being fed into the Bayesian optimizer (BO).}

\subsubsection{Microfluidic}

\added{The microfluidic device has two possible control parameters: (1) pressure to drive the inner droplet fluid (water) and (2) pressure to drive the outer suspending fluid (mineral oil), of which we know the discrete values of at all times, but do not know the combination of these values that produce optimized droplets. Both the water and oil pressure are controlled by a flow controller (Flow-EZ, Fluigent); our microfluidic device operates at pressure limits of 0-2000 mbar. Neither of these two control parameters are artificially constrained, the limits are defined by the physical limits of the hardware only. The outer mineral oil pinches off droplets of the inner DI water at a junction that is 40 \textrm{$\mu$}m wide, the droplets then travel into a channel 200 \textrm{$\mu$}m wide, shown in Fig. \ref{fig:micro_hardware}; the entire device has a height of 30 \textrm{$\mu$}m. Microfluidic droplet makers operate in one of five regimes at any given time: no inner fluid flow, unbroken inner fluid flow, dripping, transition, and jetting \cite{utada2007dripping,lignel2017water, jeyhani2019microneedle}. Our experiment explores two out of the three drop generation regimes: (1) dripping, where the drop forms while touching the constriction walls, and (2) transition, where the drop forms in the constriction without touching the walls. Jetting does not occur in our experiments due to the device structure, fluid viscosities, and experimental parameter space \cite{utada2007dripping}. Similar to the inkjet device, for every printed sample, the control parameters, normalized to the range $[0,1]$, are appended to the image and fed into the BO algorithm.

The droplet properties in the microfluidic system are primarily described by two dimensionless numbers: (1) Capillary number (Ca), which compares viscous forces to surface tension and (2) Weber number (We), which compares inertial forces to surface tension \cite{cubaud2008capillary, anna2006microscale,garstecki2005mechanism, utada2007dripping}.}
\begin{equation}
     \textrm{Ca}=\frac{\mu \dot{\gamma} a}{\sigma},\qquad
     \textrm{We}=\frac{\rho v^2 a}{\sigma},
\end{equation}
where $\mu$ is the viscosity of the outer fluid or continuous phase, $\dot{\gamma}$ is the shear rate, $a$ is the droplet diameter, $\sigma$ is the surface tension between the two fluids, $\rho$ is the density of the continuous phase, and $v$ is the characteristic flow velocity. 
\replaced{The ranges of these dimensionless numbers within our system are found to be $\textrm{Ca}\approx$ 0.005--0.05 and $\textrm{We}\approx$ 0.0005--0.003, which are further discussed in the results. Viscous stresses dominate the inertial effects in the microfluidics system because $\textrm{Re}\approx$ 0.04--0.4.
}{Our model represents the microfluidic device as a function of three control parameters: (1) pressure to drive the inner droplet fluid (water) and (2) pressure to drive the outer suspending fluid (mineral oil),which we know the discrete values of at all times. Pressure is controlled by a flow controller (Flow-EZ, Fluigent) operating over a range of 0-2000 mbar. The outer mineral oil pinches off droplets of the inner DI water at a junction that is 30 \textrm{$\mu$}m wide, the droplets then travel into a channel 200 \textrm{$\mu$}m wide; the entire device has a height of 30 \textrm{$\mu$}m. The experiment can be operating in one of five regimes: no inner fluid flow, unbroken inner fluid flow, dripping, transition, and jetting \cite{utada2007dripping}. Our experiment achieves two out of the three drop generation regimes: dripping, where the drop forms while touching the constriction walls, and transition, where the drop forms in the constriction without touching the walls. Jetting does not occur due to the device structure, fluid viscosities, and experimental parameter space. Similar to the inkjet device, for every printed sample, the normalized values of the control parameters are appended to the image and input into the BO.}

\subsection{Parameter Space Initialization}

Without \textit{a priori} knowledge of the parameter space dynamics from a domain expert, we look towards using machine learning methods, in particular, Bayesian optimization (BO), to learn the topology of the parameter space. As more experimental data is acquired, BO refines its search of the parameter space for optimized printing conditions. \added{The user collects training data over the range of parameter space in which optimization is desired. For the purpose of this demonstration, data was collected over the entire range of experimentally accessible parameters within our system.}

To give BO the best chance at finding a global optimum condition, we provide the algorithm with our best estimation of the parameter space as a whole in the form of an initialization dataset. Latin hypercube sampling (LHS) is a sampling tool that generates an initialization sample set $\mathbf{x}\in\mathcal{X}^{(N)}$ to capture the variability of an $N$-dimensional parameter space with low bias \cite{McKay2000}. LHS stratifies the range of each control parameter into $K$ strata ($K=20$ in this paper) of equal marginal probability such that each stratum is randomly sampled once to generate $\mathcal{X}^{(N)}$. The low variance demonstrated by LHS in literature, relative to random and stratified sampling, illustrates the goodness of LHS as an unbiased estimator for selecting initialization conditions \cite{McKay2000}.

\subsection{Computer Vision Characterization} \label{sec:cv}

\begin{figure}[ht]
\centering
\added[id=AES, comment={Add image of computer vision processing to clarify that the analysis is scale-invariant. Changing the hardware orifice or geometry of microchannel will still allow for optimization since the algorithm can detect and analyze droplets across multiple length-scales. Additionally, add an explanation for why stringent image capture is not required.}]{\includegraphics[width=0.75\textwidth]{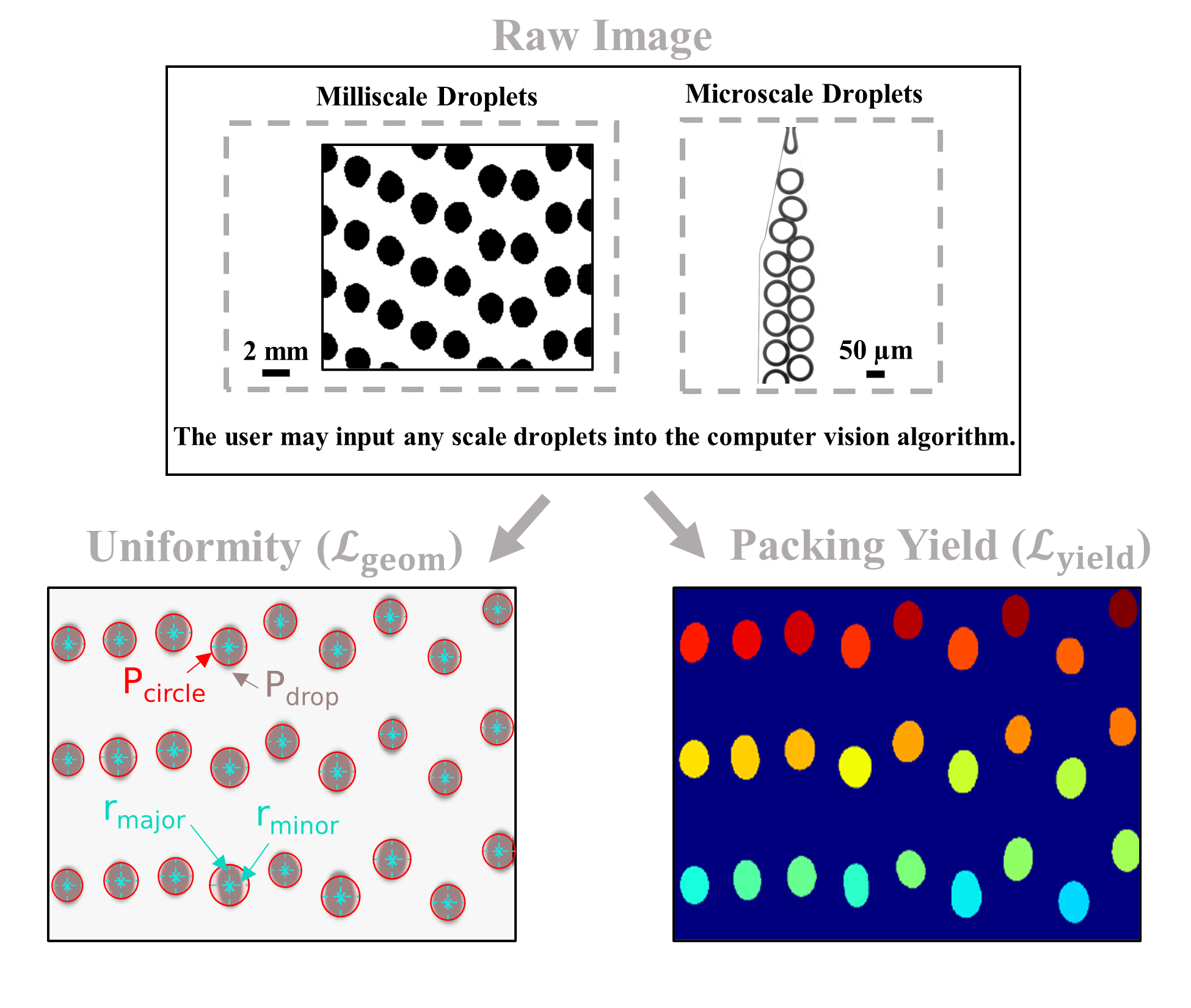}}
\caption{\added{Computer vision characterization of droplets based on circularity and yield targets. The watershed process segments the droplets from the image background such that $\mathcal{L}_\textrm{geom}$ and $\mathcal{L}_\textrm{yield}$ can be computed. The developed computer vision code is robust to variations in droplet size, allowing for optimization of droplets at several scales, without adjustment -- in this paper, milliscale and microscale droplets are explored. The developed computer vision algorithm is robust to variations in image artifacts by using pre-processing thresholds such that uniform image normalization is achieved.}}
\label{fig:cv}
\end{figure}

The user can define a target. We defined our targets to be the geometric circularity and yield of droplets\added{, as shown in Fig. \ref{fig:cv}}. An optimization objective label $\ell$ is computed for each droplet sample via computer vision, which quantifies these properties of the droplet flow. A simple scalarized linear combination of loss functions is implemented to support flexibility of the user to implement their own loss functions.

After generating each sample, the flow is imaged and fed into the computer vision software. The computer vision software utilizes a watershed segmentation process flow \cite{itseez2015opencv}. The watershed process defines a dynamic threshold to segment each droplet in the sample from the background, as shown in Fig. \ref{fig:cv}, such that we generate a set of indexed droplet pixels $P_{\textrm{drop}}$. We characterize all indexed droplets within a sample by the optimization objective label, $\ell\in[0,1]$, which we aim to minimize.

Droplet circularity is the first component of $\ell$ and it is calculated by computing the loss between all droplets in the image $P_{\textrm{drop}}$ and perfect circles mapped onto the centroids of each droplet $P_{\textrm{circle}}$.
\begin{equation} 
\label{eq:geometric}
\mathcal{L}_{\textrm{geom}} = \frac{\sum \overline{P_{\textrm{drop}} \cap P_{\textrm{circle}}}}{\sum P_{\textrm{drop}}}.
\end{equation}
By computing the major axis chord $r_{\textrm{major}}$ and minor axis chord $r_{\textrm{minor}}$ of each indexed droplet, a perfect circle is mapped to each droplet such that the circle's centroid is the intersection point of $r_{\textrm{major}}$ and $r_{\textrm{minor}}$ and the radius is the average of these chords. Fig. \ref{fig:cv} illustrates the process of mapping a circle to each watershed indexed droplet in the sample. We aim to minimize $\mathcal{L}_{\textrm{geom}} \in [0,1]$ to achieve droplets of high circularity.

Droplet yield is the second component of the optimization objective and it is computed by taking the loss of all droplet pixels and all non-droplet pixels to minimize the number of non-droplet pixels.
\begin{equation} 
\label{eq:yield}
\mathcal{L}_{\textrm{yield}} = \frac{1}{2}\frac{\sum \overline{P_{\textrm{drop}} }}{\sum P_{\textrm{drop}}\cup\overline{P_{\textrm{drop}}}} + \frac{1}{2}\frac{\max(\textbf{count})-\textbf{\textrm{count}}(P_{\textrm{drop}})}{\max(\textbf{count})},
\end{equation}
where $\max(\textbf{count})$ denotes the maximum count over all $P_{\textrm{drop}}$, which keeps $\ell$ as a minimization problem. We aim to minimize $\mathcal{L}_{\textrm{yield}} \in [0,1]$ to achieve droplets of high area and high count.

The total loss function $\ell(\mathbf{x})$ is constructed by taking the average of $\mathcal{L}_{\textrm{geom}}$ and $\mathcal{L}_{\textrm{yield}}$, as is common practice in computer vision studies \cite{Gatys2015}. We give equal importance to the geometric circularity and yield objectives during optimization which is suitable for our purposes. Every experimentally generated sample is labeled with a value of $\ell$ such that we aim to minimize $\ell \in [0,1]$ to achieve droplets of both high geometric circularity and high yield. The objective function in this study is user-defined, meaning that other important measurable droplet properties can be explicitly included by the user and tailored on the basis of the application.

\subsection{Bayesian Optimization} \label{sec:acquisition}

\begin{algorithm}[t]
\small
   \caption{\added{Droplet Optimization by BO in Loop}}
   \label{alg:bo}
\begin{algorithmic}
   \STATE {\bfseries Inputs:} \quad \textbf{(1)} LHS-initialized set $\mathbf{x}\in\mathcal{X}^{(N)}$, the $N$-dimensional control parameter space. \\\qquad \qquad \quad \textbf{(2)} Batch size, $b$.
   \STATE 
   \STATE {\bfseries Outputs:} \textbf{(1)} Vector of optimized control parameter values, $\mathbf{x}^*=\underset{\mathbf{x}\in \mathcal{X}}{\operatorname{arg\ min}}\; \ell(\mathbf{x})$. \\\qquad \qquad \, \, \textbf{(2)} Running minimum loss across all samples, $\ell^*=\textbf{\textrm{min}}(\ell(\mathbf{x}))$.
   \STATE 
   \STATE Let $\ell(\mathbf{x}) = \frac{\mathcal{L}_\textrm{geom}(\mathbf{x}) + \mathcal{L}_\textrm{yield}(\mathbf{x})}{2}$, the scoring function from computer vision.
   \STATE Let $\ell(\mathbf{x}) \in \mathcal{S}\; \forall\; \mathbf{x} \in \mathcal{X}^{(N)}$, the set of losses of the training samples.
  \WHILE{no convergence}
       \STATE \textbf{Learn:} Estimate parameter space with the GP surrogate, $\ell_\textrm{pred}(\mathbf{x})\sim\mathcal{GP}(\cdot)$.
       \STATE \textbf{Predict:} Using an acquisition function $a\in\{EI, MPI, LCB\}$, predict control parameter values, $\mathbf{x}_i$ for sample $i$, which minimize $\ell_\textrm{pred}(\mathbf{x}_i)$.
       \STATE \textbf{Synthesize:} Synthesize $b$ samples using $\mathbf{x}_i$ parameters and image the sample.
       \STATE \textbf{Update:} Update $\ell_\textrm{pred}(\mathbf{x}_i)$ with the computer vision image measurement $\ell(\mathbf{x}_i)$. Concatenate $\mathbf{x}_i$ to $\mathcal{X}^{(N)}$ and $\ell(\mathbf{x}_i)$ to $\mathcal{S}$.
  \ENDWHILE
   \STATE \textbf{return} $\mathbf{x}^*$, $\ell^*$.
\end{algorithmic}
\end{algorithm}

We drive the efficient discovery of unstable fluid conditions using BO in loop, shown in Algorithm \ref{alg:bo}, where previous batches of generated samples inform the next batch of generated samples. The control parameter values $\mathbf{x}\in\mathcal{X}^{(N)}$ and labels $\ell\in\mathcal{S}$ of previous batches serve as a likelihood to a probabilistic surrogate model, in this case a Guassian Process (GP) \cite{gpy2014}. The GP utilizes a decision policy, called an acquisition function, to acquire new control parameter values $\mathbf{x}$ that better optimize $\ell(\mathbf{x})$ or explore new regions of the parameter space that may contain more optimized values of $\ell$. We aim to minimize $\ell(\mathbf{x})$ to find the control parameter conditions that produce highly circular and high yield droplets \cite{Seeger2004,Brochu2010}:
\begin{equation}
    \mathbf{x}^*=\underset{\mathbf{x}\in \mathcal{X}}{\operatorname{arg\ min}}\; \ell(\mathbf{x}),
\end{equation}
where $\mathbf{x}$ is an $N$-dimensional vector of printing condition in the set $\mathcal{X}^{(N)}$.

GP is the surrogate model used to predict how the droplet flow $\ell(\mathbf{x})\in\mathcal{S}$ changes as the control parameter values $\mathbf{x} \in \mathcal{X}^{(N)}$ change: 
\begin{equation}
    \replaced{\ell_\textrm{pred}(\textbf{x})}{\ell(\textbf{x})} \sim \mathcal{GP}(\mu(\mathbf{x}),\kappa(\mathbf{x},\mathbf{x}')).
\end{equation}
A Gaussian prior, $\mathcal{N}$, is assigned to the initial GP likelihood of the objective function $\ell(\mathbf{x})$ to estimate its posterior mean $\mu(\mathbf{x})$ and (co)variance $\kappa(\mathbf{x},\mathbf{x}')$ \cite{Seeger2004}:
\begin{align}
    \mu(\mathbf{x}) &= \mathbb{E}[\ell(\mathbf{x})],\\
    \kappa(\mathbf{x},\mathbf{x}') &= \mathbb{E}[(\ell(\mathbf{x})-\mu(\mathbf{x}))(\ell(\mathbf{x}')-\mu(\mathbf{x}'))].
\end{align}
In this work, three common acquisition functions are used to guide the iterative selection of $\mathbf{x}$ \cite{Seeger2004,Brochu2010}: (1) expected improvement (EI), (2) maximum probability of improvement (MPI), and (3) lower confidence bound (LCB). We introduce these three distinct acquisition functions to illustrate the robustness of the computer vision-BO in loop method for discovering control parameter values that produce unstable fluid flow in both milliscale and microscale droplet-generating devices.

We begin an experiment by initializing the algorithm using LHS. Then, we iterate interleaving sampling and update steps. For every iteration of the BO in loop, 10 new $\mathbf{x}$ are acquired using the acquisition function's decision policy. The decisions of the acquisition function are based on the GP parameter space estimation of $\ell$, computed from the labeled batch set $\mathcal{X}^{(N)}$, which contains all previous batch results. These new $\mathbf{x}$ are concatenated to $\mathcal{X}^{(N)}$ and $\ell$ are concatenated to $\mathcal{S}$ before acquiring the next batch of $\mathbf{x}$. The decision policies for each acquisition function used in this study are detailed next.

\subsubsection{Expected Improvement Acquisition}

Expected improvement (EI) is a decision policy that samples optima $\mathbf{x}$ from the GP by evaluating the function $\ell(\mathbf{x})$ and comparing the value to the current running minimum function evaluation $\ell^*=\textbf{\textrm{min}}(\ell(\mathbf{x}))$. Since our batch size is 10 samples per iteration, the EI acquisition function outputs 10 samples that have the highest improvement from previous $\ell^*$ \cite{Hennig2012,Brochu2010}:
\begin{equation}
\label{eq:a_ei}
    a_\textrm{EI}(\mathbf{x})=\int^{\ell^*}_{-\infty}(\ell^*-\ell(\mathbf{x}))\mathcal{N}(\ell;\mu(\mathbf{x}),\kappa(\mathbf{x},\mathbf{x}'))\textrm{d}\ell.
\end{equation}
A local penalization evaluator is used which penalizes points closely sampled to each other within a batch, thus, the best point is selected first, then the second best (within some distance), and so on.

\subsubsection{Maximum Probability of Improvement Acquisition}

Maximum probability of improvement (MPI) is a decision policy that samples optima $\mathbf{x}$ similarly to EI, however, MPI does not scale the function values proportional to the magnitude of improvement. This means that MPI computes the location where $\ell^*=\textbf{\textrm{min}}(\ell(\mathbf{x}))$ but does consider how much the function value $\ell$ improves from the current running minimum $\ell^*$. Thus, the acquisition of MPI is the same as EI but without the $(\ell^*-\ell)$ term \cite{Hennig2012,Brochu2010}:
\begin{equation}
\label{eq:a_mpi}
    a_\textrm{MPI}(\mathbf{x})=\int^{\ell^*}_{-\infty}\mathcal{N}(\ell;\mu(\mathbf{x}),\kappa(\mathbf{x},\mathbf{x}'))\textrm{d}\ell.
\end{equation}

\subsubsection{Lower Confidence Bound Acquisition}

Lower confidence bound (LCB) is a decision policy that samples optima $\mathbf{x}$ through an explicit trade-off scheme between exploitation and exploration. A pure exploitation acquisition function outputs optima $\mathbf{x}$ exactly at evaluated GP function mean values $\mu(\mathbf{x})$ that minimize $\ell$, \textit{i.e.}, where the standard deviation $\sqrt{\kappa(\mathbf{x},\mathbf{x}')}$ about the mean is zero. A pure exploration acquisition function outputs optima $\mathbf{x}$ within some large number of standard deviations about the mean. The LCB acquisition function balances exploiting the GP mean values with exploring the GP variance by varying a hyperparameter $\beta > 0$ \cite{Brochu2010}:
\begin{equation}
\label{eq:a_lcb}
    a_\textrm{LCB}(\mathbf{x};\beta) = \mu(\mathbf{x})-\beta\sqrt{\kappa(\mathbf{x},\mathbf{x}')}.
\end{equation}

\subsection{Evaluating Parameter Convergence}

To evaluate the performance of the presented BO and computer vision method, we visualize the topology of the objective $\ell$ within the parameter space of each trial for both devices. First, density estimations of each experimentally sampled parameter are shown to illustrate the regions of the parameter space most searched by the tool for each device and each acquisition function. Next, a subset of the imaged droplet flows are shown. Finally, the measured $\ell$ are plotted within the parameter space to demonstrate the learning of the method. Parameters with low $\ell$ generate the most circular and highest yield droplet flows for both devices at both milliscale and microscale. We compute the sample efficiency of the method by determining how many sampled points fall within the \textit{feasibility bounds}. We identify that $\ell < 0.75$ demarcates conditions that can feasibly generate droplets.

\section{Results}
\subsection{Experimentally Sampled Conditions}

\begin{figure}[] 
\centering
\begin{subfigure}[b]{0.8\textwidth}  
\includegraphics[width=\textwidth]{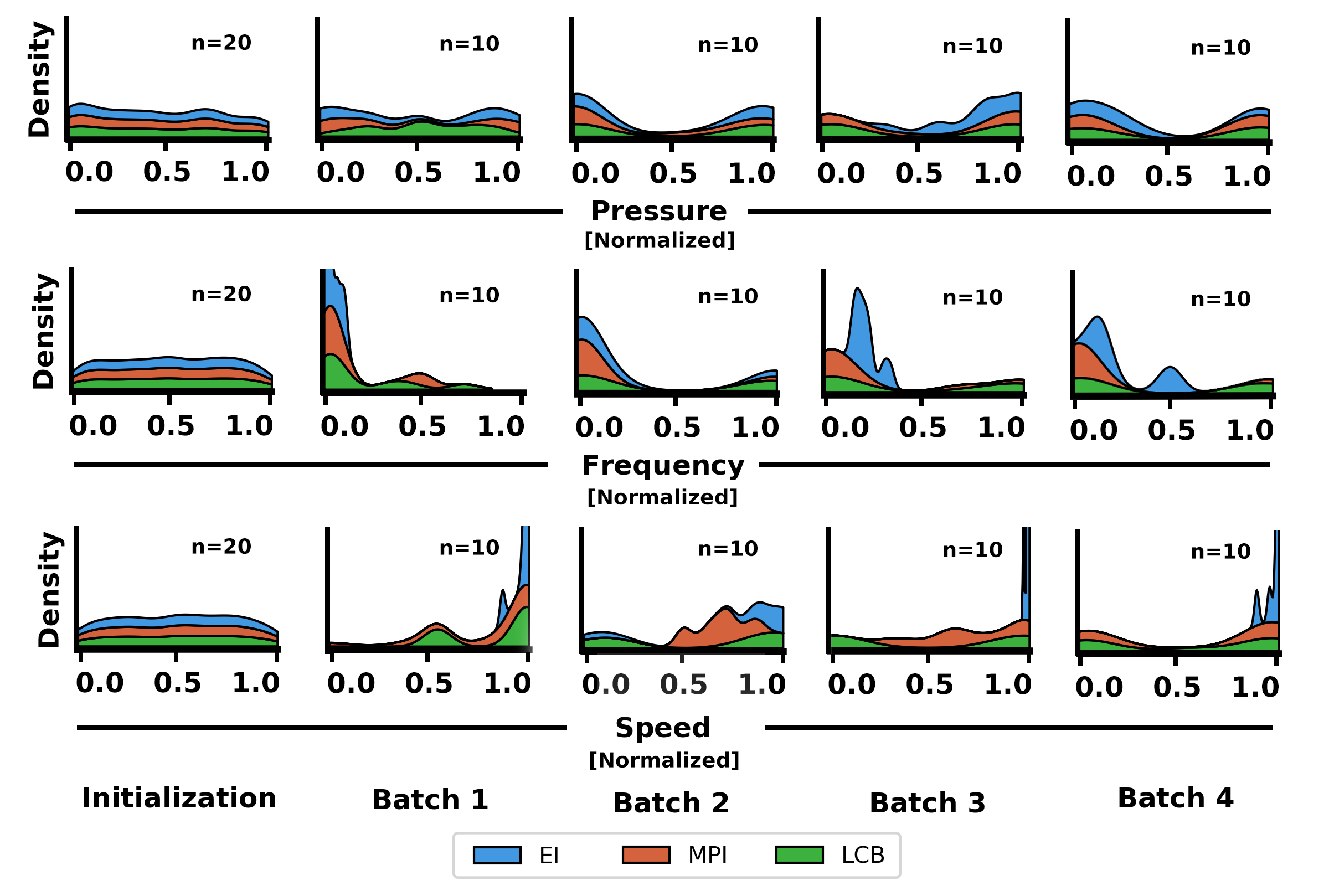}
\caption{Inkjet Control Parameter Sampling}
\label{fig:kde_inkjet}
\end{subfigure}\hfill%
\begin{subfigure}[b]{0.8\textwidth} 
\includegraphics[width=\textwidth]{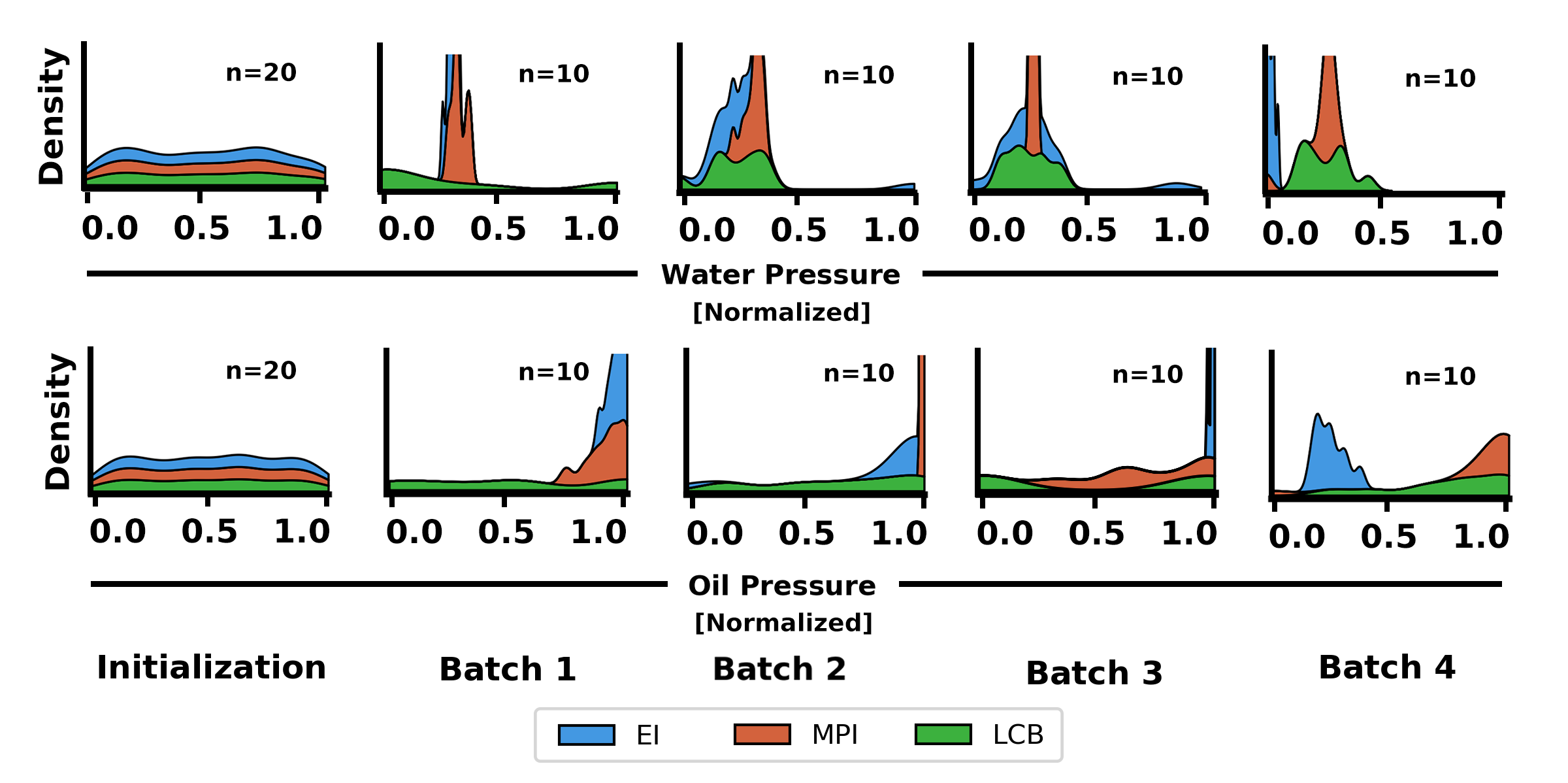}
\caption{Microfluidic Control Parameter Sampling}
\label{fig:kde_micro}
\end{subfigure}\hfill%
\caption{\comment{id=AES, comment={Adjust scale of this figure to improve readability.}}Density of control parameter values explored by each acquisition function over four batches of samples for the (a) inkjet device and (b) microfluidic device. As the number of batches increases, BO becomes more certain of what control parameter values produce the most optimized droplets.}
\label{fig:kde_flow}
\end{figure}

The computer vision-driven BO in loop approach presented in this paper is demonstrated to consistently discover optimum control parameter conditions within both milliscale and microscale droplet-generating devices. Demonstrating the optimization capabilities of this approach at two different length scales illustrates its robustness and utility as a universal tool for experimentalists to tune their droplet devices without physics-based models, regardless of length scale. \added{Furthermore, even though the phenomena of droplet breakup is governed by non-linear differential equations, our approach efficiently determines an appropriate set of experimental parameters to obtain the desired drops.}

Fig. \ref{fig:kde_inkjet} illustrates the regions of the inkjet parameter space explored by each acquisition function for each batch of conditions acquired by BO in loop. It should be noted that these curves are density estimations of sampling frequency for each parameter and, thus, interpolate the shape of the curve between points. The inkjet device is driven by control parameters of pressure, frequency, and speed such that BO predicts parameters $\mathbf{x}\in X^{(3)}$ that minimize $\ell(\mathbf{x})$. For the pressure condition, it is shown that BO discovers samples across the whole range of pressures, indicating that the droplet flows are not sensitive to changes in pressure. Conversely, BO sampling is more concentrated for the frequency and speed conditions, indicating that high circularity and high yield droplet flows form using low frequency and high speed conditions.

\begin{figure}[ht]
\centering
\begin{subfigure}[b]{0.7\textwidth}  
\includegraphics[width=\textwidth]{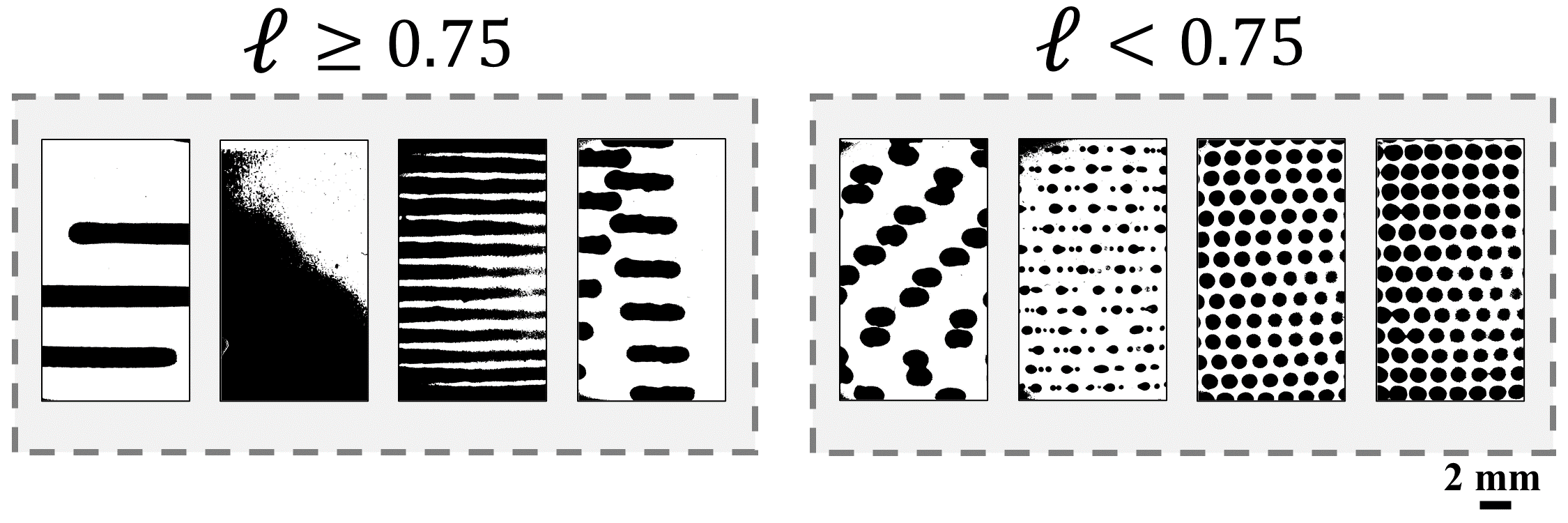}
\caption{Inkjet Droplets}
\label{fig:l_inkjet}
\end{subfigure}\hfill%
\begin{subfigure}[b]{0.68\textwidth} 
\includegraphics[width=\textwidth]{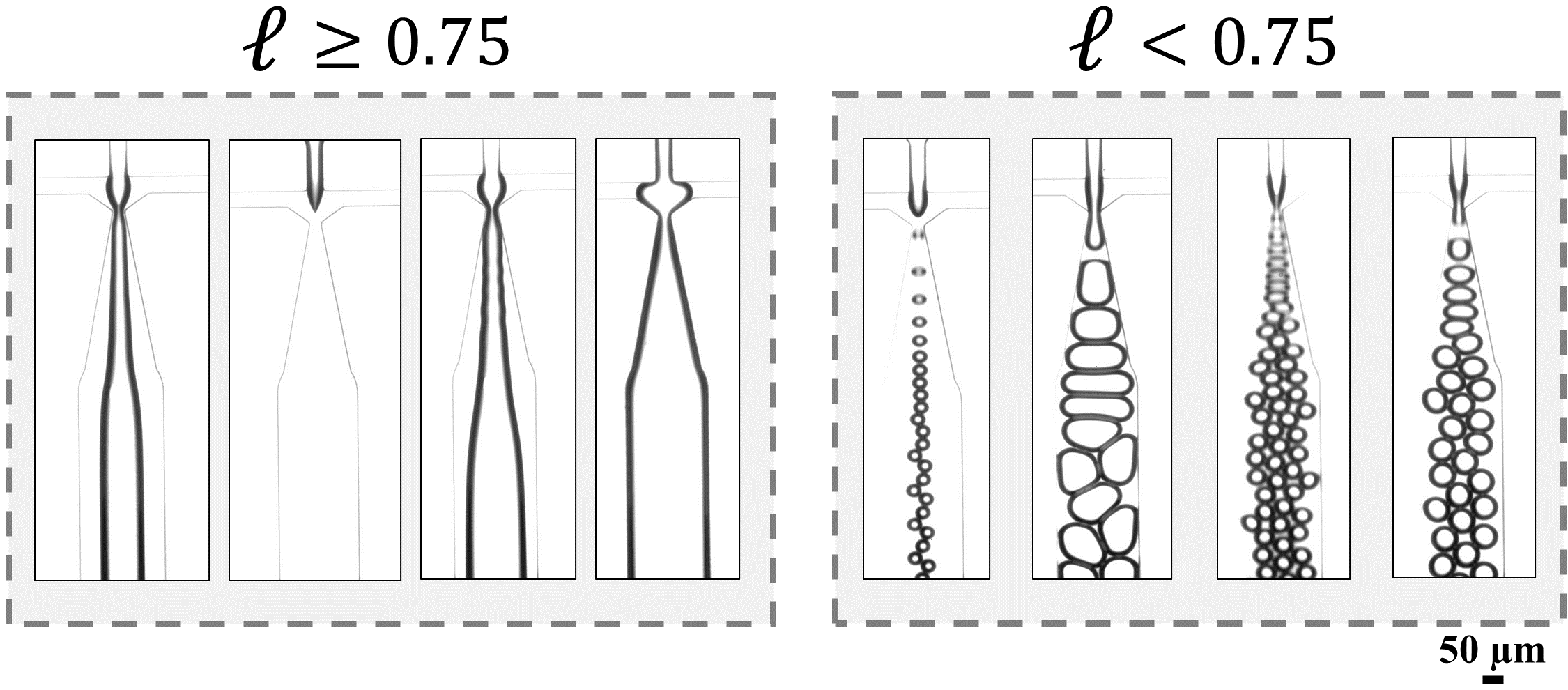}
\caption{Microfluidic Droplets}
\label{fig:l_micro}
\end{subfigure}\hfill%
\caption{Experimentally generated droplet samples labeled with the computer vision-computer optimization objective, $\ell$ for the (a) inkjet device and (b) microfluidic device. Control parameters that generate a sample with $\ell<0.75$ are shown to produce discrete droplets, hence, exhibiting the feasibility of Rayleigh and capillary instability occurring at those conditions. Conversely, control parameters that generate a sample with $\ell \geq 0.75$ are shown to have continuous or connected fluid flows, implying it is not feasible for Rayleigh or capillary instability to occur at these conditions.}
\label{fig:l}
\end{figure}

Fig. \ref{fig:kde_micro} illustrates the regions of the microfluidic parameter space explored by each acquisition function. The microfluidic device is driven by two control parameters of water pressure and oil pressure such that BO predicts parameters $\mathbf{x}\in X^{(2)}$ that minimize $\ell(\mathbf{x})$. Unlike the inkjet device, the droplet flows are shown to be sensitive to both the control parameters in the microfluidics device. BO acquires points from low water pressure conditions and high oil pressure conditions.

The results from Fig. \ref{fig:kde_flow} illustrate the sampling mechanics of the GP-EI, GP-MPI, and GP-LCB BO models when all functions are initialized using the same dataset for each respective device. The regions of the parameter space where $\ell$ is minimized are learned using the sampling mechanics of BO. It will be shown that for each experiment, all of the tested BO models converge to similar optimum control parameters.

\subsection{Learned Parameter Space Topology}

\begin{table}[]
\caption{\added{Time required to perform each process step in Bayesian optimization of droplets. Process steps are separated into whether it occurs on the software- or hardware-end of the optimization procedure. Averages and standard deviations of 12 model runs are shown, each run generating 10 droplet samples. The total time for convergence for 4 rounds of updates plus the initialization dataset is $total\ time\ per\ sample\ \textrm{[s]} * 60$, since 60 samples are generated and analyzed for an entire optimization procedure.}}
\label{tab:time}
\begin{tabular}{lll}
\textbf{Process Step} & \textbf{Time per Sample {[}s{]}} & \textbf{Type} \\ \toprule
Read Images & 0.10 $\;\;\;\pm$ 0.01 & Software \\
Computer Vision & 2.7 $\;\;\;\;\;\pm$ 0.2 & Software \\
Retrain Surrogate & 0.40 $\;\;\;\pm$ 0.1 & Software \\
Acquisition & 0.50 $\;\;\;\pm$ 0.1 & Software \\
Device Configuration & 70.0 $\;\;\;\pm$ 10 & Hardware \\
Print Droplets & 30.0 $\;\;\;\pm$ 5 & Hardware \\
Image Droplets & 35.0 $\;\;\;\pm$ 10 & Hardware \\ \hline
\multicolumn{1}{|l}{Total} & {3.7 $\;\;\;\;\;\pm$ 0.2} & \multicolumn{1}{l|}{Software} \\
\multicolumn{1}{|l}{Total} & {135.0 $\;\,\pm$ 15} & \multicolumn{1}{l|}{Hardware} \\
\multicolumn{1}{|l}{\textbf{Total}} & \textbf{138.7 $\,\pm$ 15} &  \multicolumn{1}{l|}{\textbf{Both}}\\ \hline
 &  &  \\
 & \multicolumn{2}{l}{\textbf{Time for Convergence {[}hr{]}}} \\ \hline
\multicolumn{1}{|l}{Total} & {0.062 $\;\,\pm$ 0.0004} & \multicolumn{1}{l|}{Software} \\
\multicolumn{1}{|l}{Total} & {2.25 $\;\;\;\,\pm$ 0.03} & \multicolumn{1}{l|}{Hardware} \\
\multicolumn{1}{|l}{\textbf{Total}} & {\textbf{2.31 $\;\,\,\,\pm$ 0.03}} & \multicolumn{1}{l|}{\textbf{Both}} \\
\hline
\end{tabular}
\end{table}

\begin{figure}[ht] 
\centering
\begin{subfigure}[b]{0.5\textwidth}  
\includegraphics[width=\textwidth]{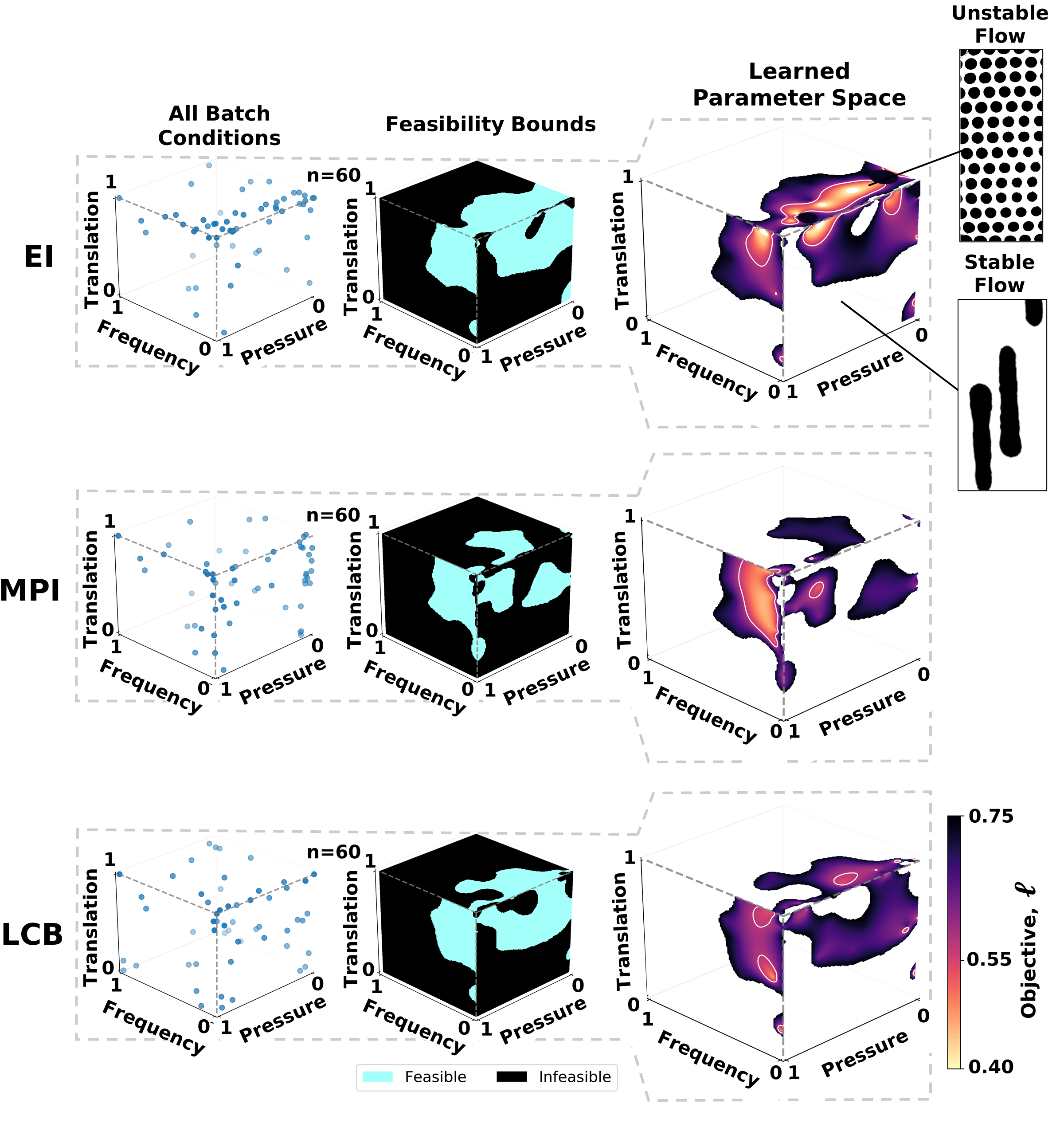}
\caption{Learned Rayleigh Unstable Conditions}
\label{fig:space_inkjet}
\end{subfigure}\hfill%
\begin{subfigure}[b]{0.5\textwidth} 
\includegraphics[width=\textwidth]{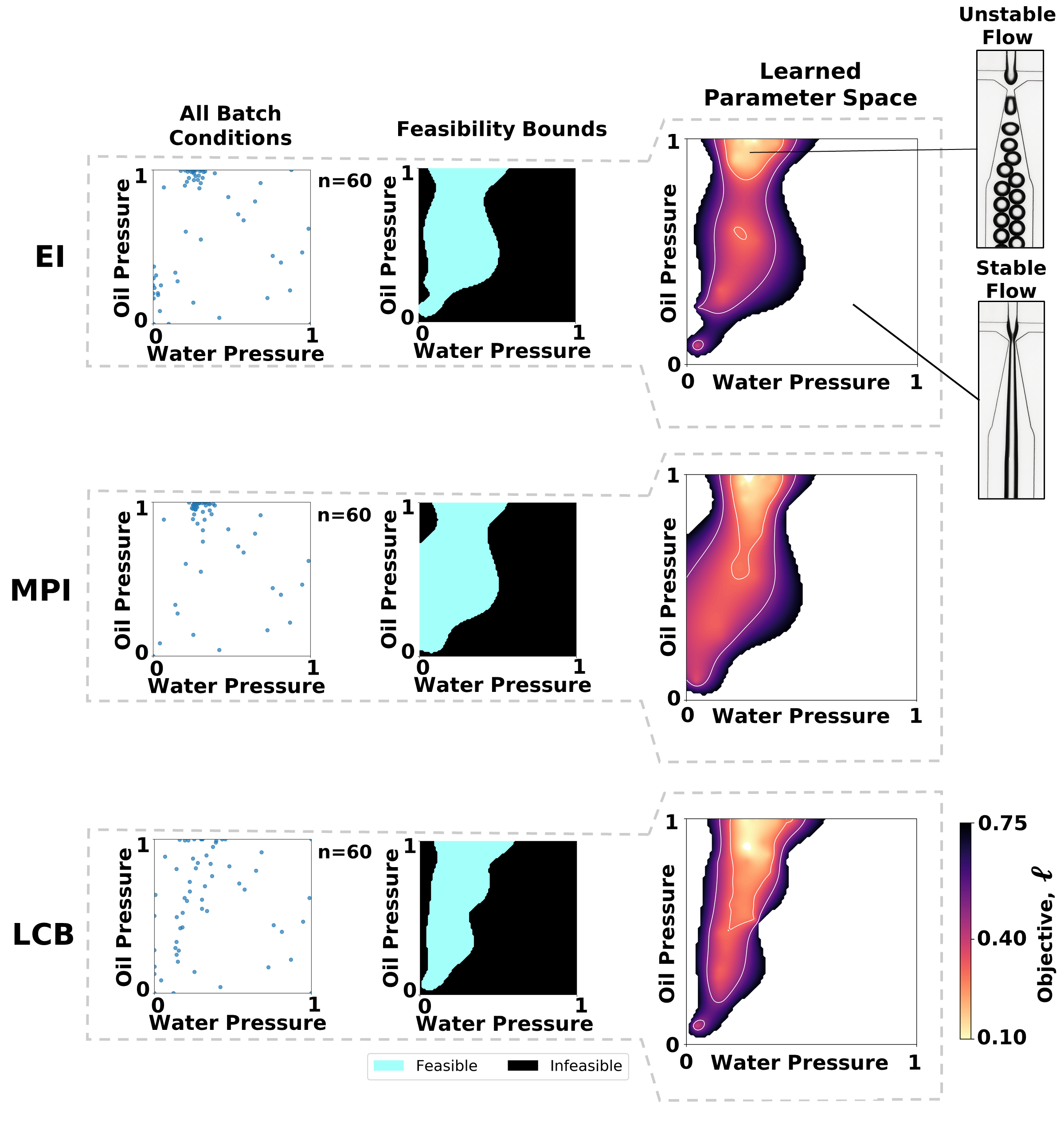}
\caption{Learned Capillary Unstable Conditions}
\label{fig:space_micro}
\end{subfigure}\hfill%
\caption{\comment{id=AES, comment={Increased figure font size.}}Control parameter values learned by BO to form (a) Rayleigh and (b) capillary instability to generate optimized droplets within the inkjet device and microfluidics device, respectively. Three rows of figures are illustrated, one row for each independent experiment run with a different acquisition function (EI, MPI, and LCB). The first column illustrates the values of all control parameter values sampled by all batches of BO over the course of the experiment. The second column illustrates the where BO has discovered that Rayleigh/capillary instability can occur within the parameter space ($\ell<0.75$). The third column shows the \added{measured} value of $\ell$ within the feasibility bounds. Lower values of $\ell$ denote more optimized droplet flows.}
\label{fig:space}
\end{figure}

Our method discovers the control parameter conditions that produce droplets due to fluid instabilities at both the millimeter and micrometer scales using the same computer vision-driven BO in loop method. \added{Table \ref{tab:time} notes the time required to perform each process step. Our method takes $138.7$ seconds per sample to run, and $2.3$ hours to run the full optimization procedure, compared to $>70$ hours via prior methods \cite{Chu2019}. The time for convergence is shown to be dominated by hardware calibration and synthesis processes (97.3\% of total convergence time), rather than software processing or model training (2.7\% of total convergence time). This is true since the learning process occurs in the loop and only $O(10^1)$ data points are used in our GP approach, compared to the $O(10^6)$ data points used for upfront training in prior neural network approaches \cite{Chu2019}.} Fig. \ref{fig:l} illustrates a subset of droplet flows from stable flows, \textit{i.e., outside of the feasibility bounds, $\ell \geq 0.75$}, and from unstable flows, \textit{i.e., inside the feasibility bounds $\ell < 0.75$}. Fig. \ref{fig:space} shows the parameter space topology of the objective values $\ell$ obtained from each experiment. It should be noted that topologies are Gaussian interpolated from the raw data for clarity; no extrapolation is done. The sample efficiency of each BO model is determined based on the number of sampled points that fall within the feasibility bounds -- meaning that a model with higher sample efficiency requires fewer samples to achieve an optimum solution.

\begin{figure}[] 
\centering
\begin{subfigure}[b]{0.5\textwidth}  
\includegraphics[width=\textwidth]{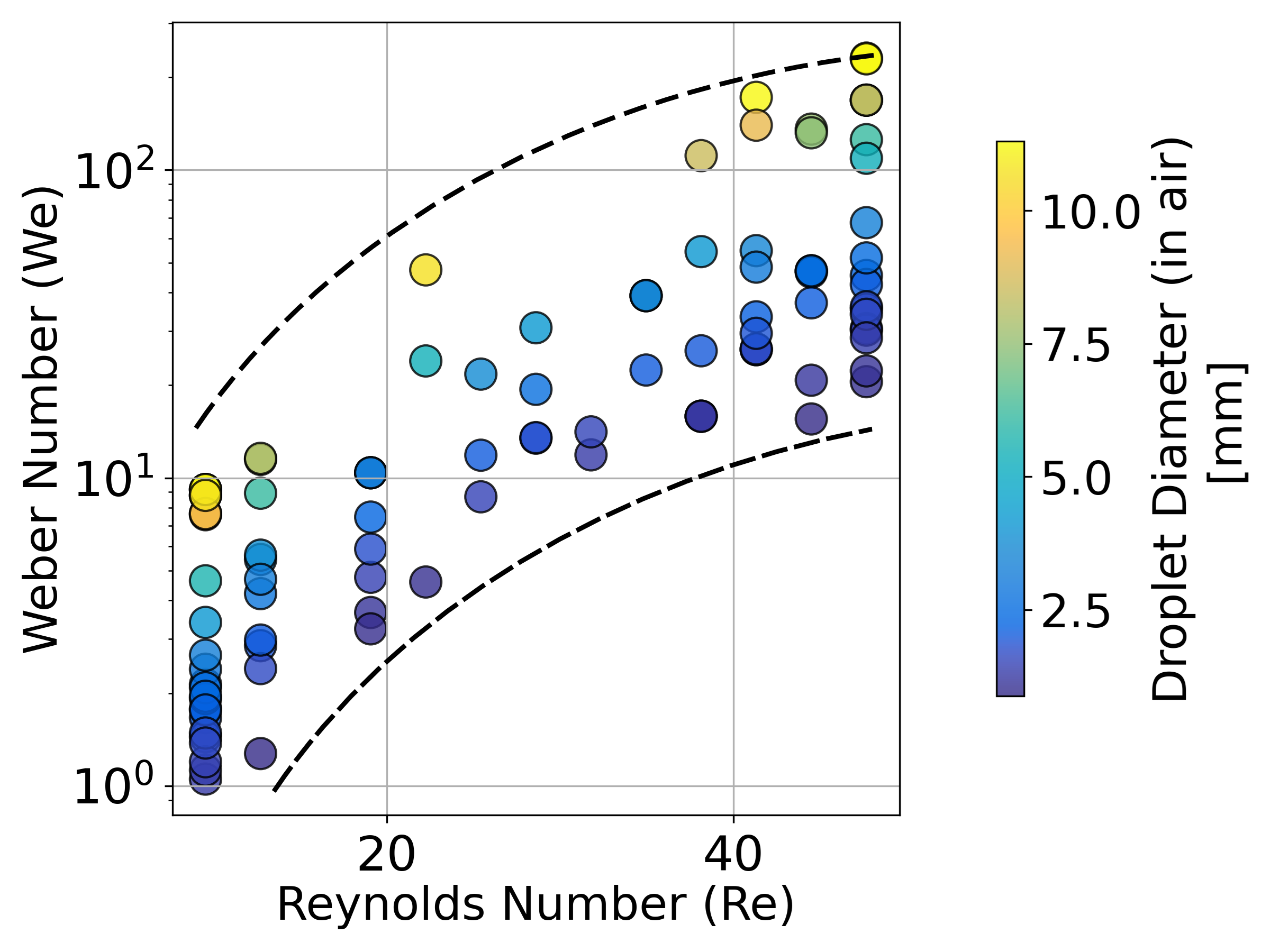}
\caption{Inkjet Drop Diameter}
\label{fig:ink_diam}
\end{subfigure}\hfill%
\begin{subfigure}[b]{0.47\textwidth} 
\includegraphics[width=\textwidth]{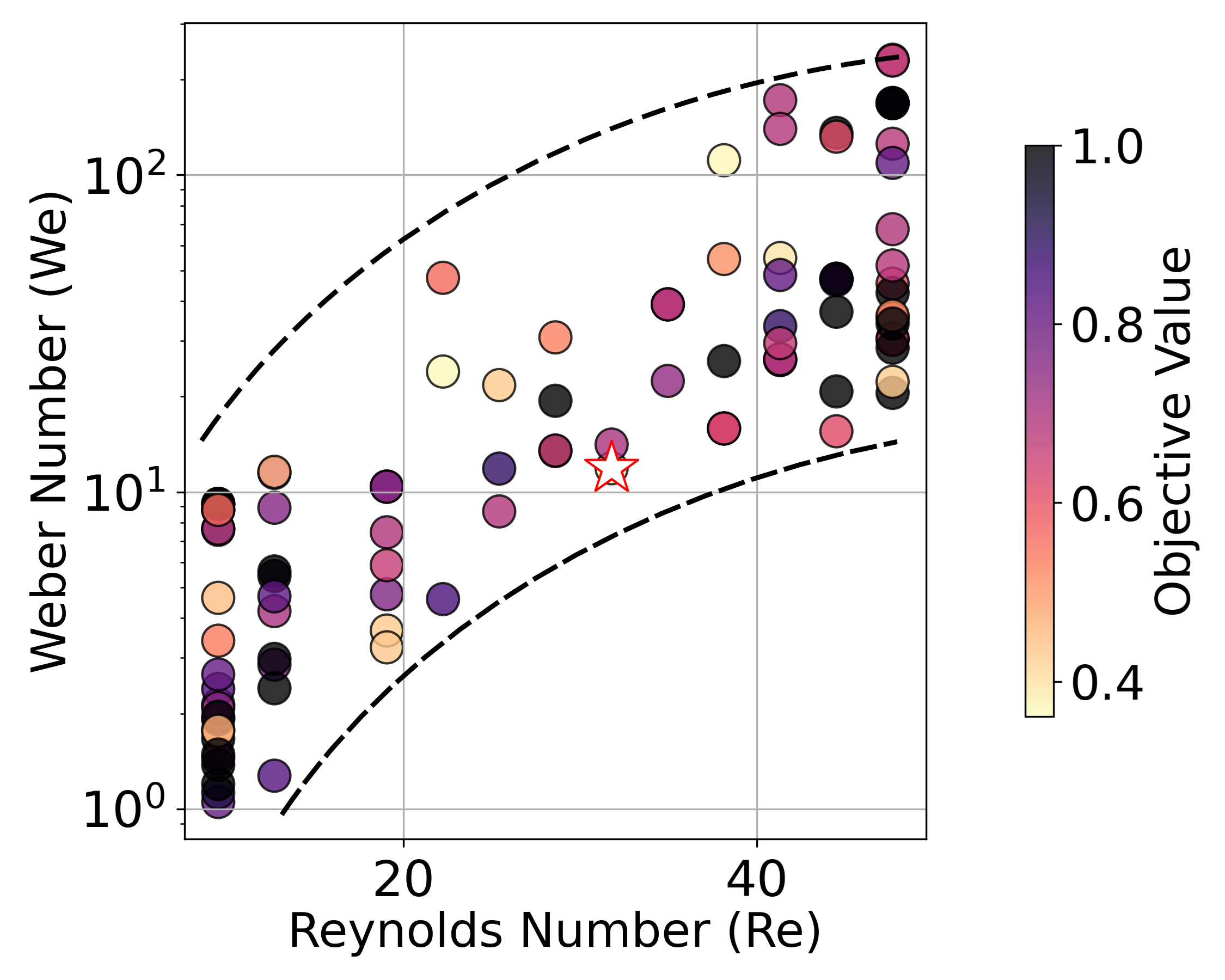}
\caption{Inkjet Objective Value}
\label{fig:ink_obj}
\end{subfigure}\hfill%

\begin{subfigure}[b]{0.47\textwidth}  
\includegraphics[width=\textwidth]{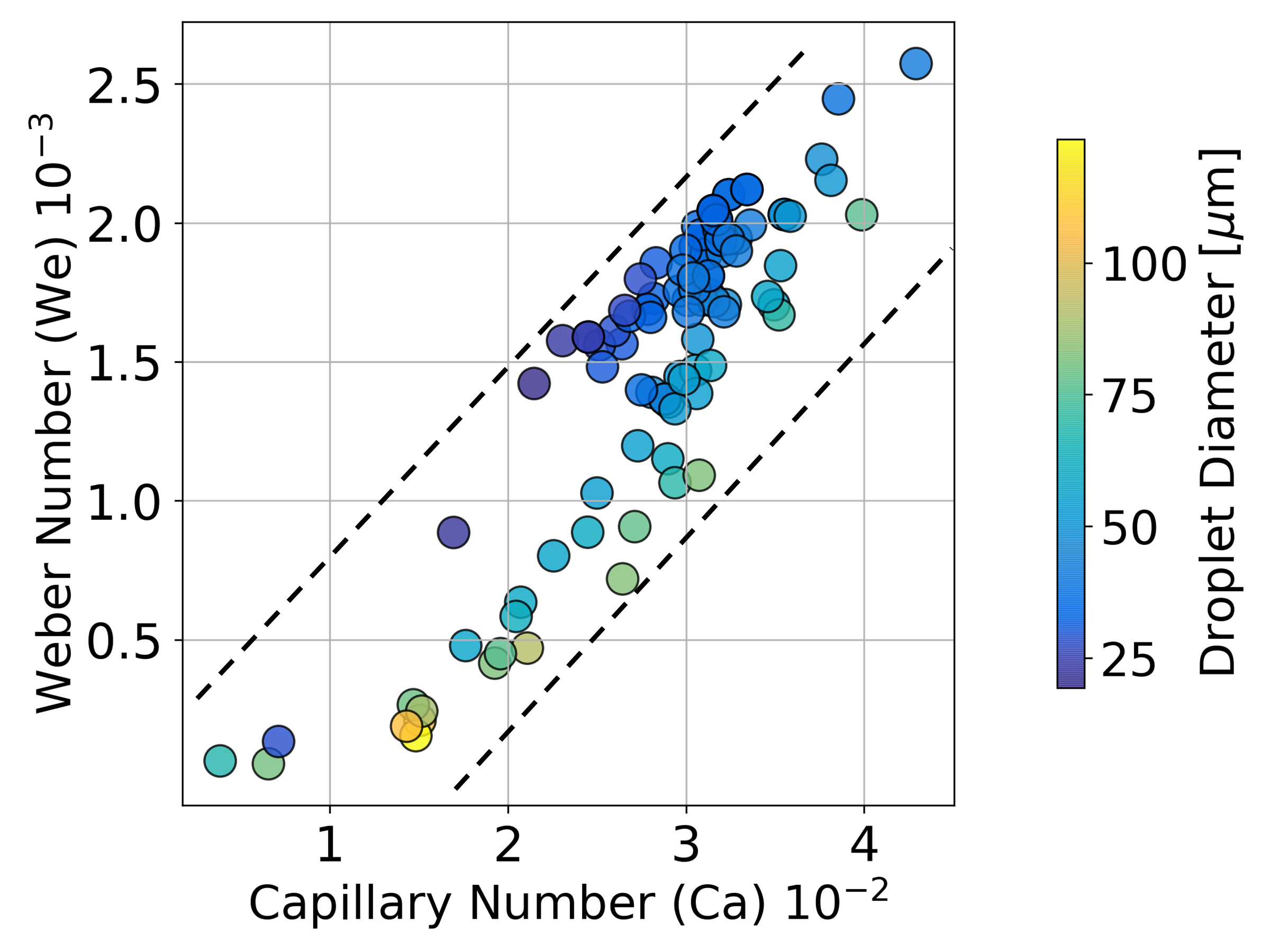}
\caption{Microfluidic Drop Diameter}
\label{fig:micro_diam}
\end{subfigure}\hfill%
\begin{subfigure}[b]{0.47\textwidth} 
\includegraphics[width=\textwidth]{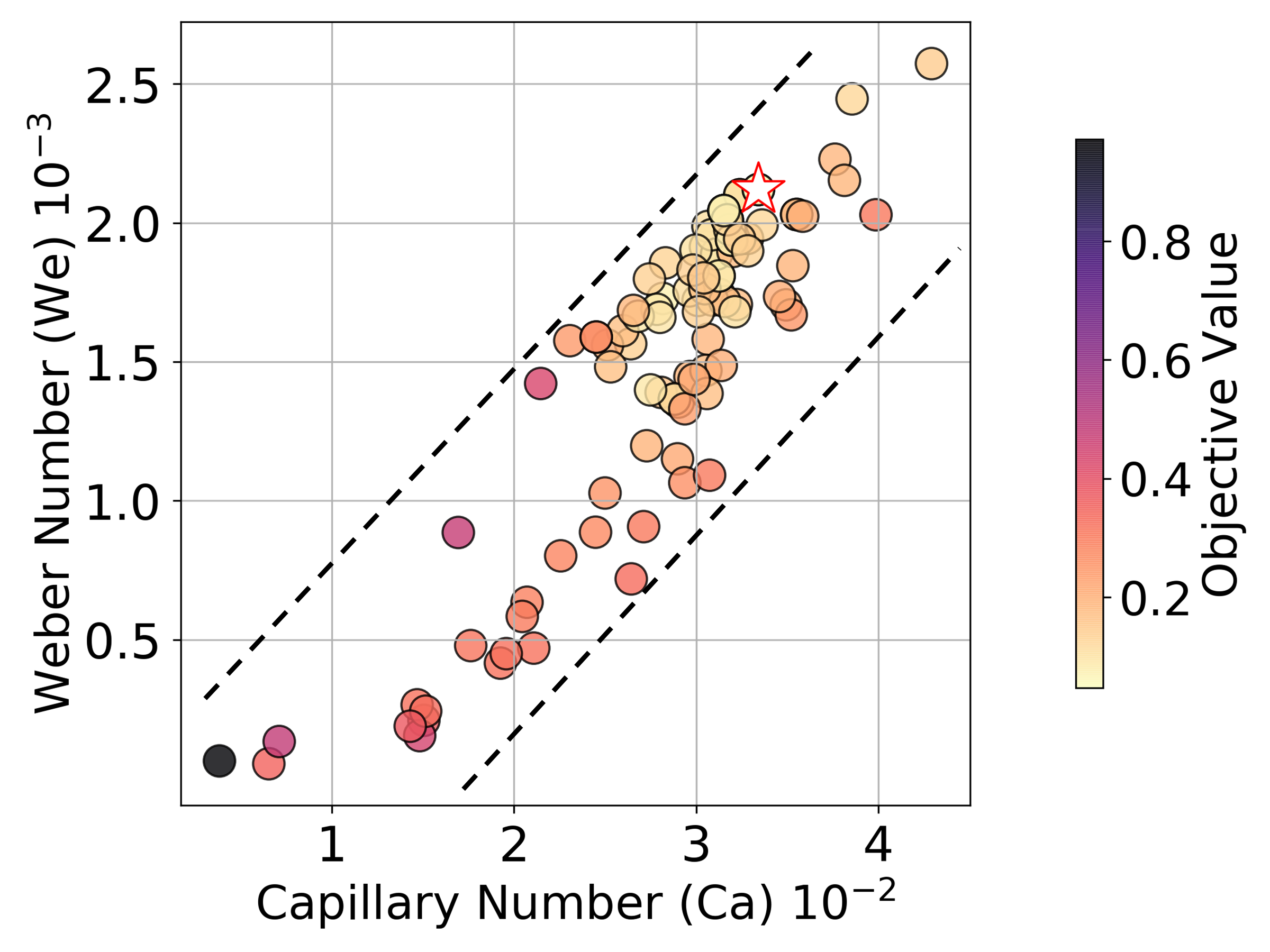}
\caption{Microfluidic Objective Value}
\label{fig:micro_obj}
\end{subfigure}\hfill%
\caption{\added{Reynolds number versus Weber number for (a) inkjet drop diameter, (b) inkjet objective value, $\ell$; and capillary number versus Weber number for (c) microfluidic drop diameter, (d) microfluidic objective value, $\ell$. Each subfigure includes all $n=120$ synthesized droplet patterns across the three acquisition policy experiments, not including the initialization set. The dashed curves represent the envelop of the parameter space explored. For the inkjet device, the diameter of the in-air droplets are computed using a spherical end-cap and the measured contact angle \cite{Sommers2008}. For the microfluidics device, the velocity and fluid density are referring to the continues phase (mineral oil) and the interfacial tensions for water in oil. The location of the most optimal droplet pattern, \textit{i.e.,} lowest objective value, is indicated as a white star in (b) and (d).}}
\label{fig:dimensionless}
\end{figure}

For the inkjet device, BO discovers the feasible control parameter values for Rayleigh instability to be in low-mid range frequencies and mid-high range translation speeds using different decision policy acquisition functions. Moreover, Fig. \ref{fig:space_inkjet} illustrates that this instability may exist for all pressure values within the inkjet device. The importance ranking of each parameter-objective relationship is indicated using SHAP in Fig. \ref{sfig:shap-ink} for the inkjet device \cite{shap2017}. The model sample efficiencies when integrated into the optimization of the inkjet device are: (1) 65.0\% using GP-EI, (2) 45.0\% using GP-MPI, and (3) 40.0\% using GP-LCB. \added{Fig. \ref{fig:ink_obj} shows a scatter plot of $\textrm{Re}$ versus $\textrm{We}$ for every inkjet-printed sample output by the BO model; the colorbar indicates the objective value. These experimental results illustrate the highly non-linear relationship between $\textrm{Re}$, $\textrm{We}$, and the objective value. Despite the complexity of this parameter space, our method is still shown to discover the correct optimized region of the parameter space as three independent experimental trials converge to the same optimum conditions, as shown in Fig. \ref{fig:space}. These optimized conditions fall within the regions of the $\textrm{Re}$--$\textrm{We}$ space that have optimal drop-on-demand printing conditions, as suggested in the literature \cite{McKinley2011,Derby2010}. Our results in Fig. \ref{fig:ink_obj} validate the expectation that droplets do not reliably form due to Rayeligh instability at $\textrm{We}\lesssim 3$ \cite{Derby2010}. The computed We may differ slightly from the expected value since the jet diameter contracts and expands as a function of velocity \cite{Eggers2008}. Comparing these results to Fig. \ref{fig:space_inkjet} and Fig. \ref{sfig:inkjet_acq}, the optimized parameter space is shown to correspond to low actuation frequency and fast translation speed throughout the pressure range.} Additional information regarding the impact of control parameters on droplet optimization can be found in the Supplementary Information.

For the microfluidic device, BO discovers the feasible control parameter values for capillary instability to be in low-mid range water pressure values for the whole range of oil pressure values using different decision policy acquisition functions, as shown in Fig. \ref{fig:space_micro}. The importance ranking of each parameter-objective relationship is indicated using SHAP in Fig. \ref{sfig:shap-micro} for the microfluidic device \cite{shap2017}. The model sample efficiencies when integrated into the optimization of the microfluidic device are: (1) 67.5\% using GP-EI, (2) 97.5\% using GP-MPI, and (3) 75.0\% using GP-LCB. \added{Predictions of droplet shape and size as a function of the device control parameters remain intractable to predict \cite{baroud2010dynamics}. Fig. \ref{fig:micro_diam}--\ref{fig:micro_obj} illustrate the non-linear relationship between $\textrm{We}$, $\textrm{Ca}$, and the objective value, further demonstrating the significance of exploring the experimental configuration space via BO. The optimized objective value is found in the region $\textrm{Ca}\sim$ 0.03 and $\textrm{We}\sim$ 0.002. Comparing these results to the optimization shown in Fig. \ref{fig:space_micro} and Fig. \ref{sfig:micro_acq}, a roughly linear band, in which the oil and water pressures increase proportionally to each other, can be seen. However, the non-linear bounds on the learned optimized region in the parameter space, shown again in Fig. \ref{fig:space_micro}, are a reminder of the non-linear physics governing droplet formation and characteristics.}

\section{Discussion}

Our work addresses the challenge of developing a universal tool for optimizing droplet-generating devices at multiple length scales. It is often the case where no analytical function exists to map device control parameters to a target material property, thus, resulting in wasteful trial-and-error experimentation to discover those optimized conditions. \replaced{We demonstrate the ability of a computer vision-integrated Bayesian optimization framework to accurately and rapidly discover control parameter values with repeatability that optimize the generated droplets with no \textit{a prioi} domain knowledge of the device.}{We demonstrate the ability of a computer vision-integrated Bayesian optimization framework to accurately and repeatedly discover control parameter values that optimize the generated droplets with no \textit{a prioi} domain knowledge of the device.} The significance of this paper is in providing a scale-invariant and physical model-agnostic method for droplet-generating devices optimization \added{that can be used to flexibly optimize many different droplet-generating systems in $2.3$ hours}.

We demonstrate that for these milliscale and microscale droplet-generating devices, a machine learning algorithm can learn which control parmeter values produce Rayleigh/capillary instabilities and, in turn, transition a continuous fluid stream to a decomposed flow of droplets. The performance of this method is demonstrated using three independent experiments for each device, each with a different acquisition function but with the same initialization dataset and surrogate model: (1) GP-EI, (2) GP-MPI, and (3) GP-LCB. After four batches of data sampling and model updates, all three experiments converge to similar optimized regions of the parameter space for both length scale devices. \added{We illustrate the significance of our method in an application space where droplets optimized for high circularity and high yield do not always correspond with droplets of a certain size, see Fig. \ref{fig:ink_diam}--\ref{fig:ink_obj}, hence, demonstrating the power of a flexible ML method to predict the values of control parameters that optimize droplets. Convergence occurs within four rounds of sampling, resulting in a total average optimization time of $2.3$ hours, compared to $>70$ hours previously reported using supervised learning, hence, accelerating throughput by $30\times$ \cite{Chu2019}. Unlike previous literature where the majority of the optimization time is dominated by model training, the majority of optimization time using our method is dominated by running experiments on the device hardware. Therefore, this method provides an extremely fast way to optimize droplets within complex parameter spaces and has the potential to be accelerated further by an increase in experimental throughput.}

For the inkjet device, converged conditions are the full pressure range, low piezoelectric frequencies, and high axial translation speeds. For the microfluidic device, converged conditions are low-mid water pressures and high oil pressures. However, each of these acquisition functions has a different sampling efficiency, indicating that the acquisition function affects the rate of learning differently depending on the length scale of the droplets generated by a device. The highest efficiency acquisition function for the inkjet device is GP-EI where 65.0\% of sampled points are within the feasiblity bounds, whereas the highest efficiency acquisition function for the microfluidic device is GP-MPI where 97.5\% of sampled points are within the feasibility bounds. The differences in these results can be explained by each device having different governing physics and a different number of control parameters that comprise the parameter space. Although GP-LCB results in a low sampling efficiency, it provides a powerful mode of exploration rather than exploitation, which could be used to improve our understanding of the governing physics of droplet formation across length scales. It should be noted that even though the acquisition function was varied in the computer vision-Bayesian optimization algorithm, all experiments results in convergence to optimized droplet flows.

\section{Conclusion}

We present a machine learning-based method for \added{rapid} optimization of several droplet-generating devices at different length scales. Tuning an experimental device to produce an optimized product is dynamic; the optimized conditions change based on the operating length scale of a device due to the governing physics of the device changing. Thus, there is the challenge to design a universal method for scale-invariant droplet optimization. In this study, we \replaced{develop a method}{use one method} of Bayesian optimization in loop with computer vision characterization to discover the optimum conditions for generating droplets at two different length scale experimental devices \added{in $2$ hours, without upfront model training}.

Through several trials of experimentation, we demonstrate that our machine learning and computer vision approach discovers optimized droplet-generating device control parameter values at both the milliscale and microscale. Furthermore, our method does not require knowledge of droplet size dependence on control parameters, thus, this optimization approach is demonstrated to be scale-invariant. As no physical or mathematical model is required to perform optimization, this optimization approach is also demonstrated to be physical model-agnostic; meaning that this approach can be extended to explore more complicated devices like those using non-Newtonian fluids, polymeric fluids that crosslink in flow, or phase-change materials with few adjustments. \added{Moreover, our method implements probabilistic surrogate model training and retraining in parallel with data acquisition, significantly reducing the time required to run the optimization procedure compared to methods of upfront model training.} Developing this multiscale droplet optimization method provides researchers with a universal tool for \added{accelerating} smart device conditioning and reduces wasteful trial-and-error experimentation.

\begin{acknowledgement}

We thank James Serdy (MIT) for knowledge contributions and engineering expertise in hardware development and running physical experiments. We thank Dr. Armi Tiihonen for guidance and assistance with Bayesian optimization of physical systems. We thank Dr. Shijing Sun for contributions to methodology development and framing of this paper. This material is based upon work primarily supported by the Center to Center (C2C) International Collaboration on Advanced Photovoltaics as part of the Engineering Research Center Program of the National Science Foundation and the Office of Energy Efficiency and Renewable Energy of the Department of Energy under NSF Cooperative Agreement No. EEC‐1041895. Iddo Drori would like to thank Google for a cloud research grant.
\end{acknowledgement}

\begin{description}
\item \added{\textbf{Data Availability.} All data and code are available at \url{https://github.com/PV-Lab/ML-Multiscale-Droplets}}

\item \added{\textbf{Conflict of Interest Statement.} Author T.B. holds equity in a start-up company (Xinterra) focused on commercializing machine learning technologies for accelerated materials development.}

\item \added{\textbf{Supporting Information.} Additional information provided for the parameter space search of each decision policy, importance ranking of control parameters, relevant dimensionless numbers, and images of the experimental setups, including the data collected for all experiments.}
\end{description}

\bibliography{bibliography}
\begin{description}
\item \textbf{\added[id=AES, comment={Moved from below abstract to the end of references.}]{GitHub: \url{https://github.com/PV-Lab/ML-Multiscale-Droplets}}}
\end{description}

\newpage
\section{\added[id=AES, comment={Add table of contents figure at end of manuscript}]{For Table of Contents only}}
\begin{figure}[] 
\centering
\includegraphics[width=3.25in]{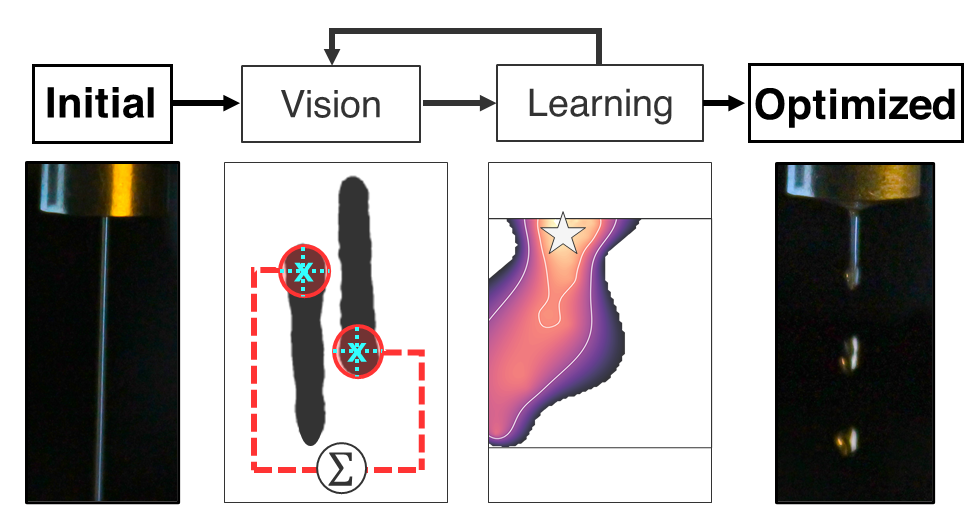}
\end{figure}

\newpage
\beginsupplement
\begin{centering}
\section{\added[id=AES, comment={Add supporting information coverpage with title, authors, affiliations, emails, and updated page numbers.}]{Supporting Information:\\A Machine Learning and Computer Vision Approach to Rapidly Optimize Multiscale Droplet Generation}}
\end{centering}

\begin{description}
\begin{centering}

\item Alexander E. Siemenn,$^{*,\dagger}$ Evyatar Shaulsky,$^{\ddagger}$ Matthew Beveridge,$^{\mathparagraph}$ Tonio Buonassisi,$^{\dagger}$ Sara M. Hashmi,$^{*,\ddagger}$ and Iddo Drori$^{*,\mathparagraph}$
\item ${\dagger}$Department of Mechanical Engineering, Massachusetts Institute of Technology, Cambridge, MA 02139, USA
\item ${\ddagger}$Department of Chemical Engineering, Northeastern University, Boston, MA 02115, USA
\item ${\mathparagraph}$Department of Electrical Engineering and Computer Science, Massachusetts Institute of Technology, Cambridge, MA 02139, USA
\item E-mail: asiemenn@mit.edu; s.hashmi@northeastern.edu; idrori@mit.edu

\end{centering}
\end{description}
\newpage

\begin{figure} 
\centering

\begin{subfigure}[b]{0.8\textwidth}  
\includegraphics[width=\textwidth]{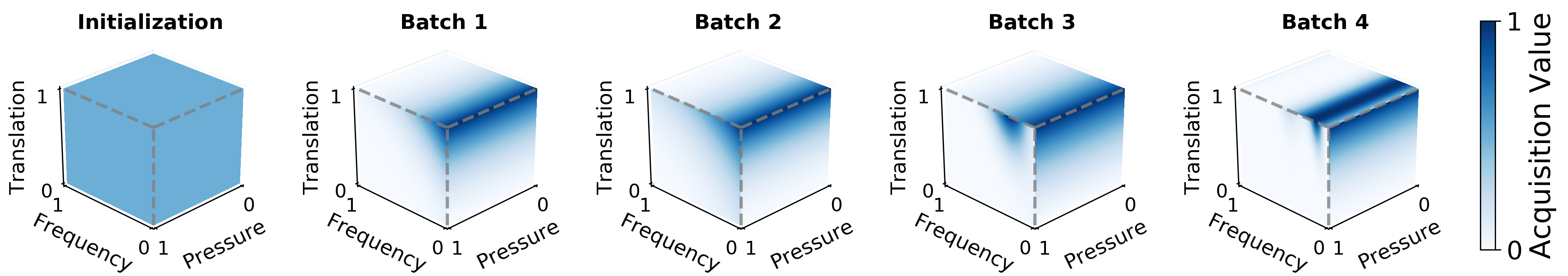}
\caption{Inkjet EI Acquisition Value per Batch}
\end{subfigure}\hfill%
\begin{subfigure}[b]{0.7\textwidth} 
\includegraphics[width=\textwidth]{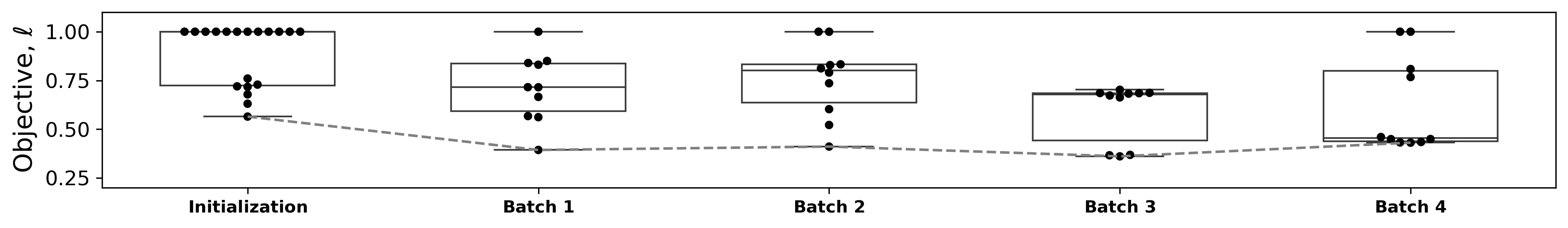}
\caption{Inkjet EI Objective Value per Batch}
\end{subfigure}\hfill%

\begin{subfigure}[b]{0.8\textwidth}  
\includegraphics[width=\textwidth]{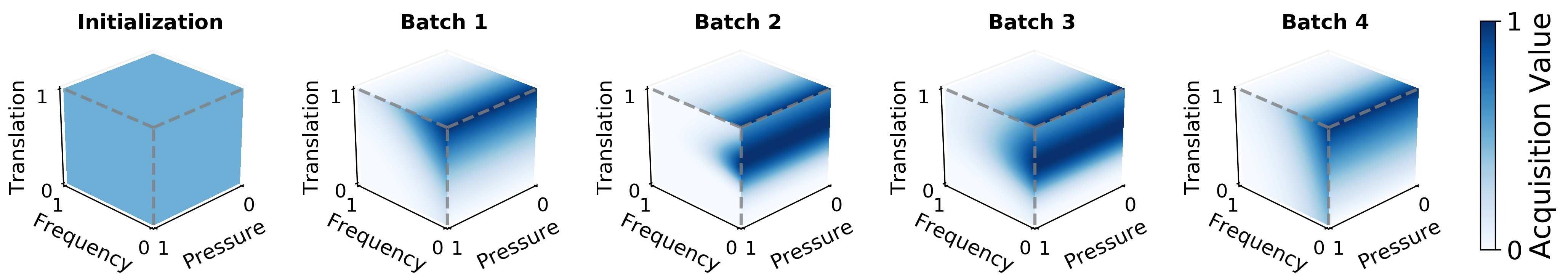}
\caption{Inkjet MPI Acquisition Value per Batch}
\end{subfigure}\hfill%
\begin{subfigure}[b]{0.7\textwidth} 
\includegraphics[width=\textwidth]{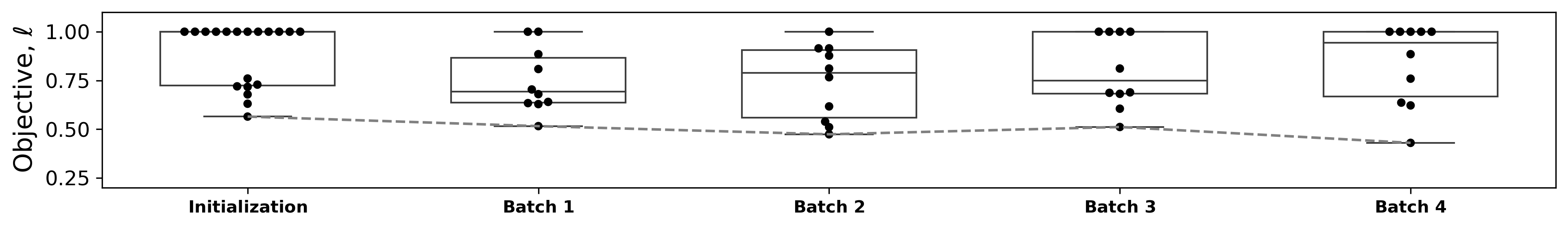}
\caption{Inkjet MPI Objective Value per Batch}
\end{subfigure}\hfill%

\begin{subfigure}[b]{0.8\textwidth}  
\includegraphics[width=\textwidth]{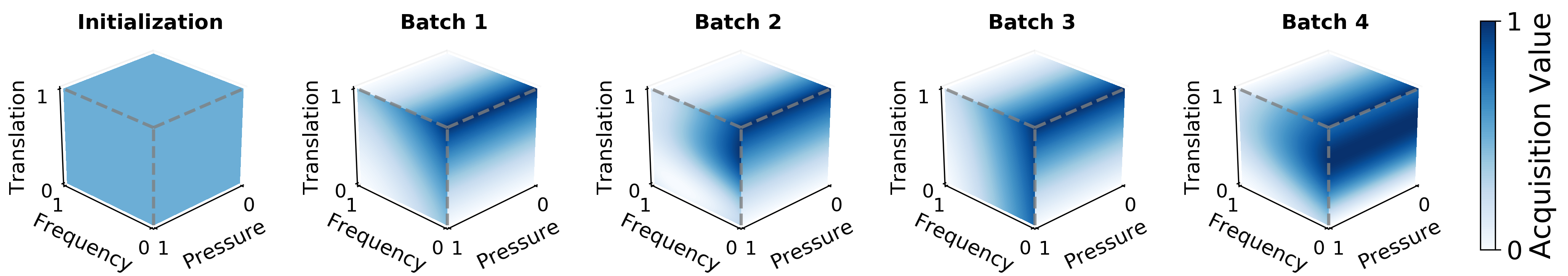}
\caption{Inkjet LCB Acquisition Value per Batch}
\end{subfigure}\hfill%
\begin{subfigure}[b]{0.7\textwidth} 
\includegraphics[width=\textwidth]{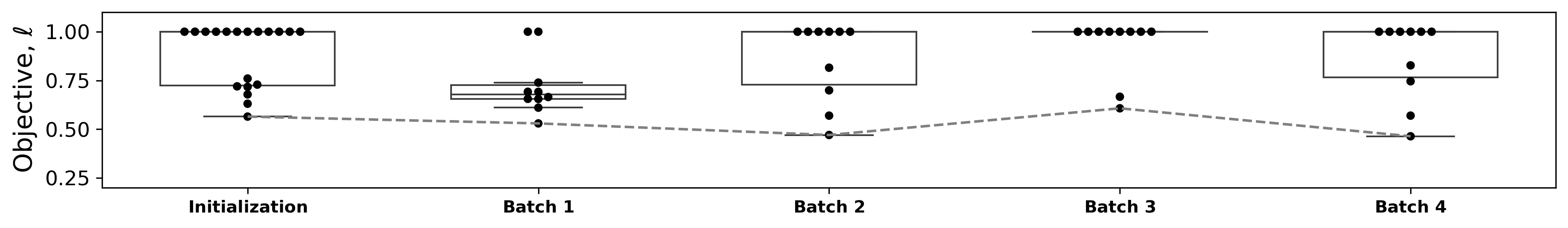}
\caption{Inkjet LCB Objective Value per Batch}
\end{subfigure}\hfill%
\caption{Acquisition function search spaces by batch in 3-dimensions. The regions where each acquisition function is likely to draw points from within the inkjet parameter space are illustrated in blue for each acquisition function: (a) EI, (c) MPI, (e) LCB -- these denote regions of high acquisition value. The initialization acquisition space is a uniform color because points are picked uniformly during initialization via LHS. Based on the mechanics of each acquisition function, the acquisition value is computed by balancing the exploitation of the GP posterior mean and exploring the GP (co)variance. The acquisition search space evolves every loop, which updates the sampling locations of data points, shown for each acquisition function: (b) EI, (d) MPI, (f) LCB. The reason why the same data point that maximizes the acquisition value is not sampled 10 times for every batch in the loop is because of evaluator used is local penalization -- points sampled close to each other are penalized, thus, the best point is selected first, then the second best (within some distance), and so on.}
\label{sfig:inkjet_acq}
\end{figure}

\begin{figure} 
\centering

\begin{subfigure}[b]{0.75\textwidth}  
\includegraphics[width=\textwidth]{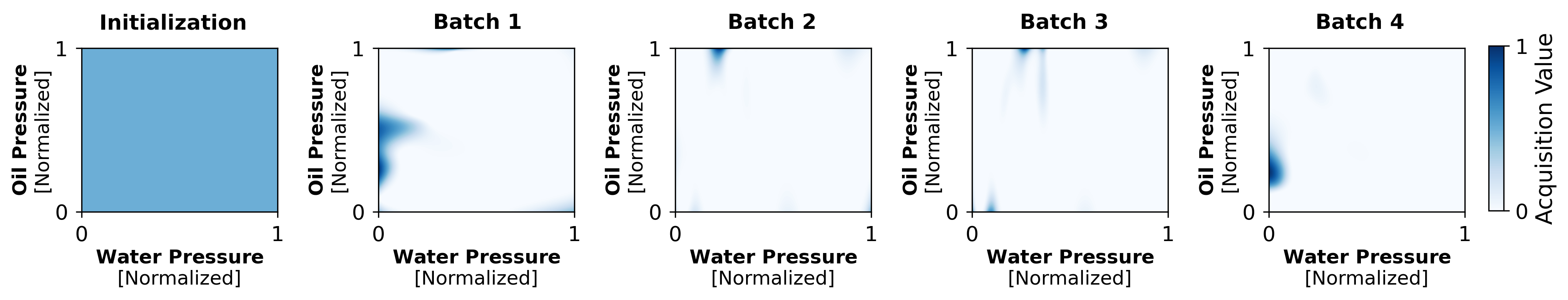}
\caption{Microfluidic EI Acquisition Value per Batch}
\end{subfigure}\hfill%
\begin{subfigure}[b]{0.67\textwidth} 
\includegraphics[width=\textwidth]{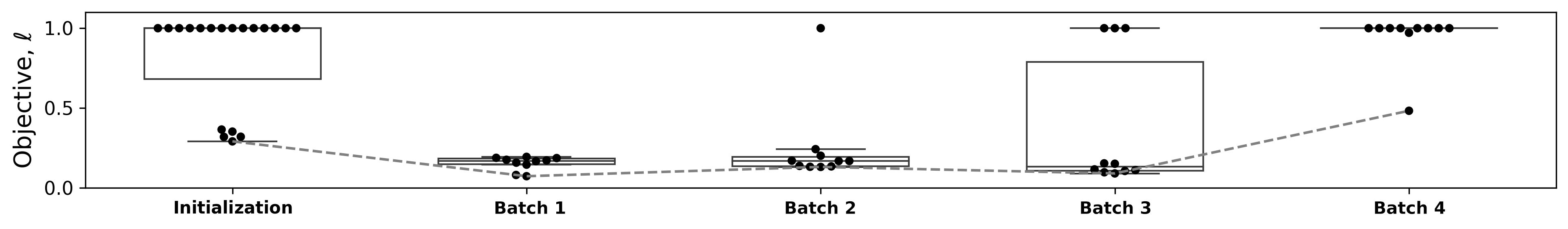}
\caption{Microfluidic EI Objective Value per Batch}
\end{subfigure}\hfill%

\begin{subfigure}[b]{0.75\textwidth}  
\includegraphics[width=\textwidth]{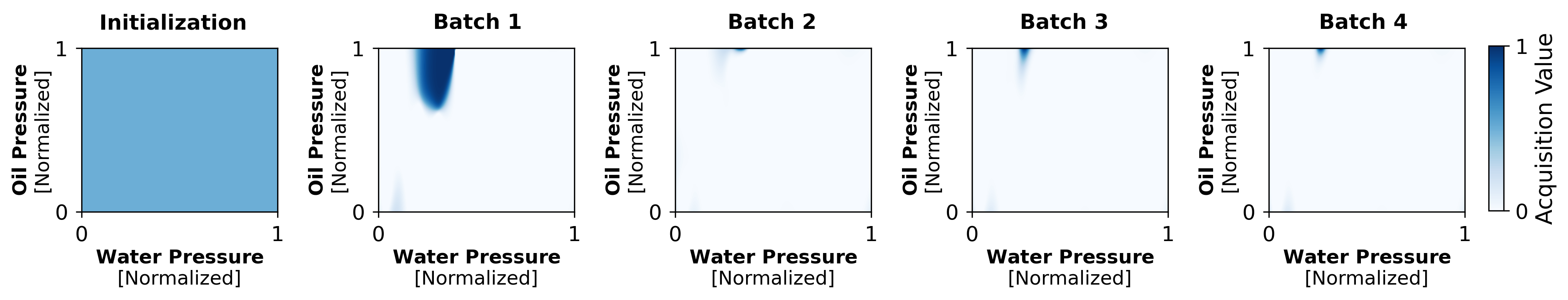}
\caption{Microfluidic MPI Acquisition Value per Batch}
\end{subfigure}\hfill%
\begin{subfigure}[b]{0.67\textwidth} 
\includegraphics[width=\textwidth]{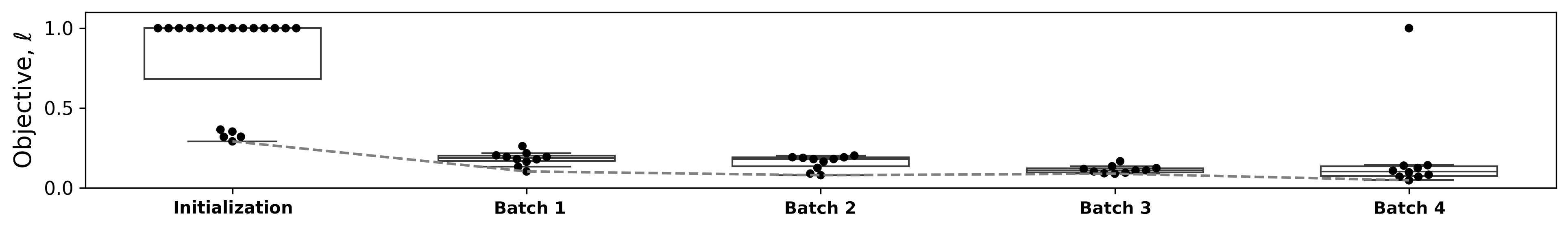}
\caption{Microfluidic MPI Objective Value per Batch}
\end{subfigure}\hfill%

\begin{subfigure}[b]{0.75\textwidth}  
\includegraphics[width=\textwidth]{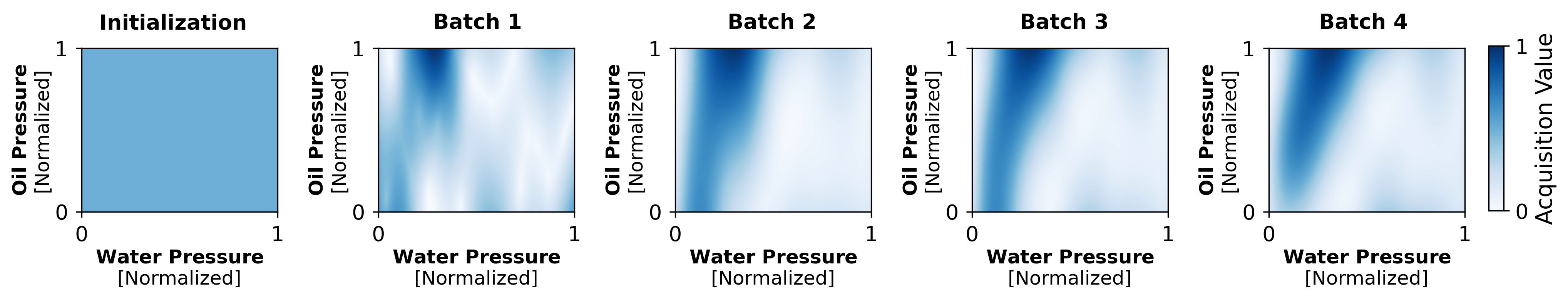}
\caption{Microfluidic LCB Acquisition Value per Batch}
\end{subfigure}\hfill%
\begin{subfigure}[b]{0.67\textwidth} 
\includegraphics[width=\textwidth]{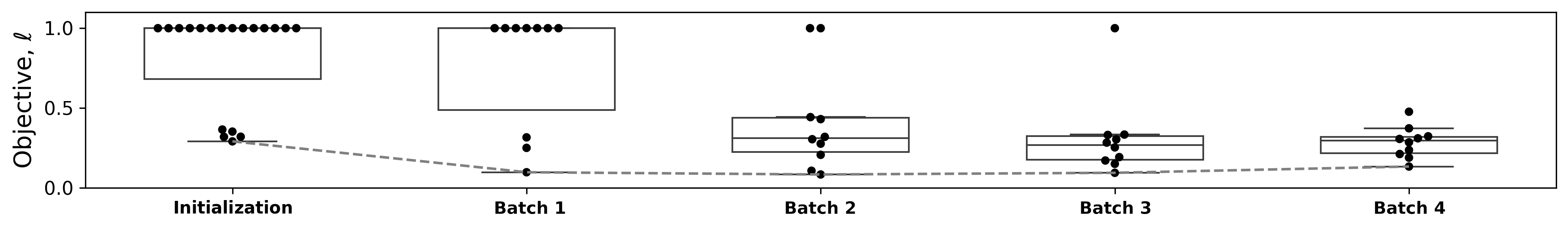}
\caption{Microfluidic LCB Objective Value per Batch}
\end{subfigure}\hfill%
\caption{Acquisition function search spaces by batch in 2-dimensions. The regions where each acquisition function is likely to draw points from within the microfluidic parameter space are illustrated in blue for each acquisition function: (a) EI, (c) MPI, (e) LCB -- these denote regions of high acquisition value. The initialization acquisition space is a uniform color because points are picked uniformly during initialization via LHS. Based on the mechanics of each acquisition function, the acquisition value is computed by balancing the exploitation of the GP posterior mean and exploring the GP (co)variance. The acquisition search space evolves every loop, which updates the sampling locations of data points, shown for each acquisition function: (b) EI, (d) MPI, (f) LCB. The reason why the same data point that maximizes the acquisition value is not sampled 10 times for every batch in the loop is because of evaluator used is local penalization -- points sampled close to each other are penalized, thus, the best point is selected first, then the second best (within some distance), and so on.}
\label{sfig:micro_acq}
\end{figure}

For each loop of optimization, the regions of the parameter space where new optima are acquired by each decision policy are highlighted in dark blue in Figure \ref{sfig:inkjet_acq} for the inkjet device and in Figure \ref{sfig:micro_acq} for the microfluidic device. For each of these sampled batches, all of the acquired points are plotted as box plots to illustrate the ranges of objective values. This information is useful when designing a fluid device such that a researcher may engineer the device to have experimental parameter values within the regions of high acquisition value.

\begin{figure}
\centering
\begin{subfigure}[b]{0.5\textwidth}  
\includegraphics[width=\textwidth]{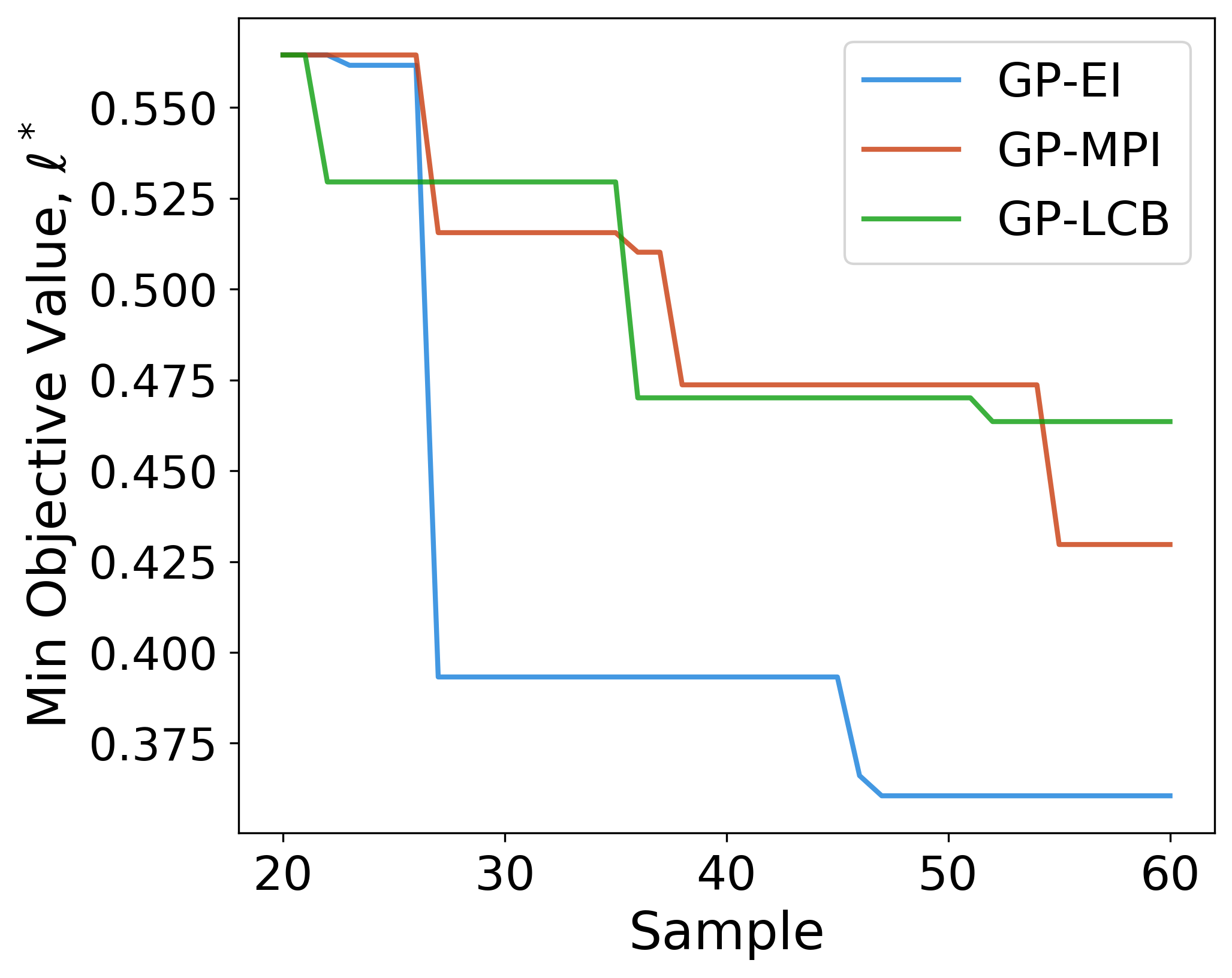}
\caption{Inkjet Running Best Objective Value, $\ell^*$}
\end{subfigure}\hfill%
\begin{subfigure}[b]{0.5\textwidth} 
\includegraphics[width=\textwidth]{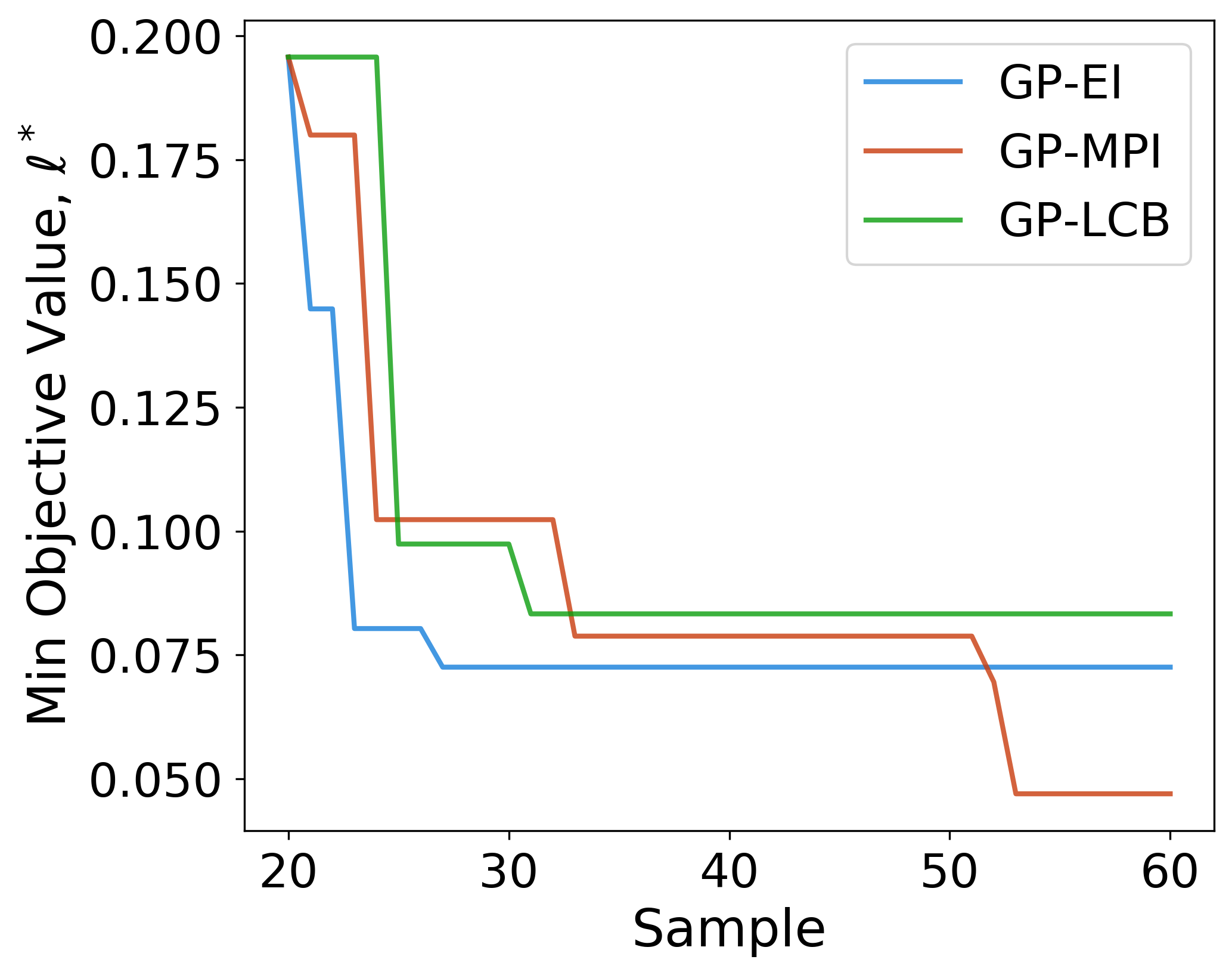}
\caption{Microfluidic Running Best Objective Value, $\ell^*$}
\end{subfigure}\hfill%
\caption{The running minimum objective function evaluation $\ell^*$ for (a) the inkjet device and (b) the microfluidic device. The optimum unstable fluid conditions that generate high yield and high uniformity droplets are labeled by $\ell^*\in[0,1]$ -- the running minimum objective value. $\ell^*$ as a function of sample number is illustrated for each acquisition function after the 20 LHS initialization data points. The vertical axes are scaled differently for the inkjet and microfluidics devices, this scaling is arbitrary and is a result of the different droplet generating dynamics within each parameter space.}
\label{sfig:run_min}
\end{figure}

Figure \ref{sfig:run_min} illustrates the running minimum objective value across all sampled points for the inkjet and microfluidic devices. As the autonomous optimization procedure runs, more highly-optimized droplet structures are discovered.

\begin{figure}
\centering
\begin{subfigure}[b]{1\textwidth} 
\includegraphics[width=\textwidth]{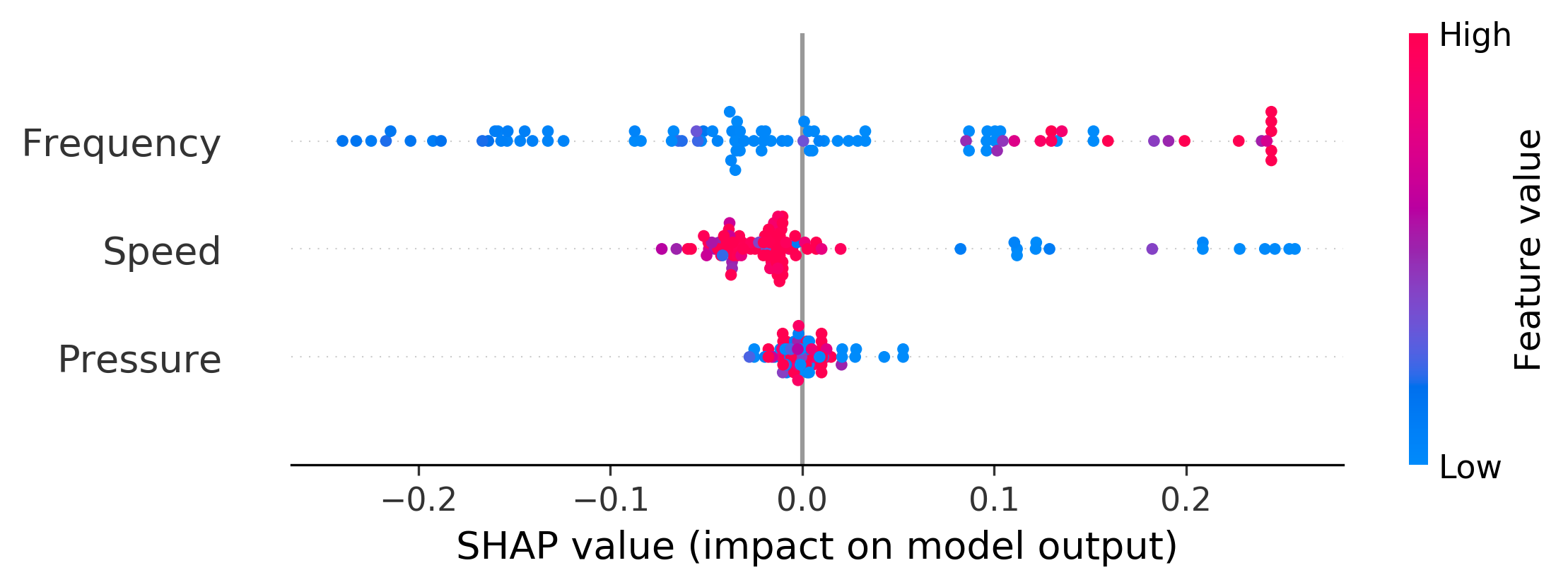}
\caption{Inkjet SHAP Feature Importance}
\label{sfig:shap-ink}
\end{subfigure}\hfill%
\begin{subfigure}[b]{1\textwidth} 
\includegraphics[width=\textwidth]{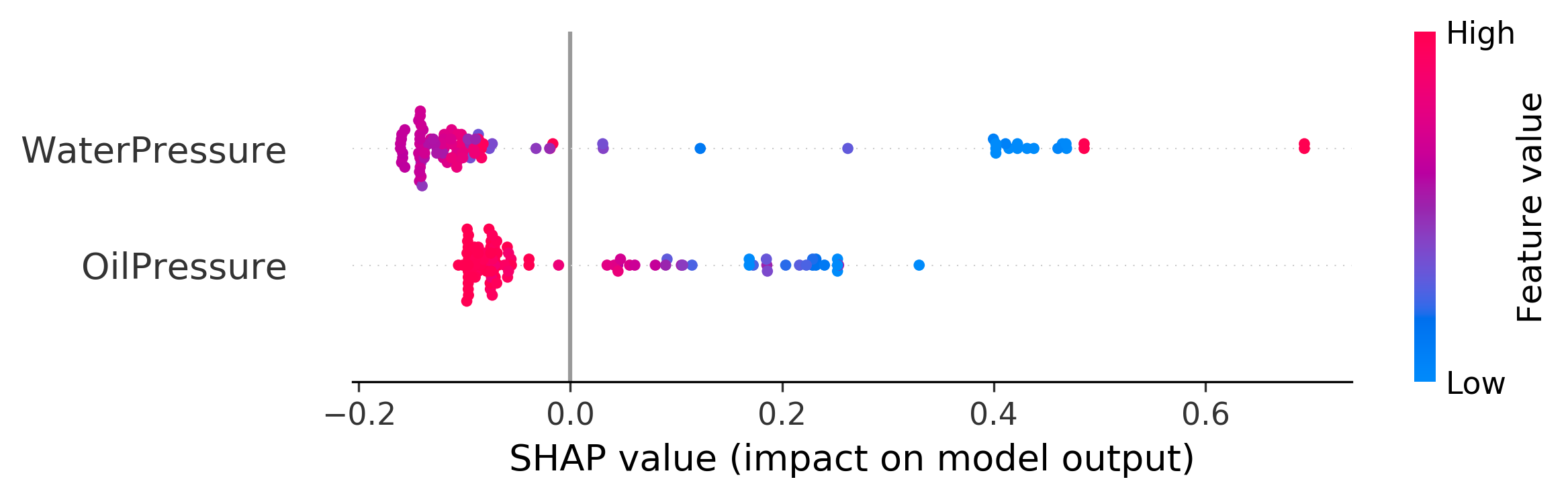}
\caption{Microfluidic SHAP Feature Importance}
\label{sfig:shap-micro}
\end{subfigure}\hfill%
\caption{The SHAP feature importance of the (a) inkjet and (b) microfluidic devices \cite{shap2017}. Random forest regression with an 80/20 training/test split is used to compute the SHAP feature importance of all acquisition function-sampled data. Each printing condition is shown to have a certain SHAP value -- the magnitude of the SHAP value describes how much that condition impacts the objective values $\ell$. For example, low frequency values (blue) are shown to generate low $\ell$ values (high yield, uniform droplets), \textit{i.e.}, illustrated by high negative SHAP values; high frequency values (red) are shown to generate high $\ell$ values (low yield, not uniform droplets), \textit{i.e.}, illustrated by high positive SHAP values. High speed values have little impact on $\ell$, whereas low speed values generate high $\ell$ values. Pressure, in general, has little impact on $\ell$, shown by the clustering of all values of pressure conditions centered on a 0.0 SHAP value.}
\label{sfig:shap}
\end{figure}

The importance impact of each experimental parameter on the objective value $\ell$ is illustrated in Figure \ref{sfig:shap}. For the inkjet device, frequency is illustrated to have a much larger impact on the objective value compared to pressure and translation speed. For the microfluidic device, water pressure is illustrated to have only a marginally larger impact on the objective valued compared to the water pressure. These results are useful for designing new fluid devices since understanding the impact of a device's input on the output product provides insight into which input parameters should receive more attention during tuning.

\deleted{An additional design condition is presented in Figure \ref{fig:dimensionless} for the microfluidic device, auxiliary to the design conditions presented in Figure \ref{fig:space}. These design conditions illustrate the (a) objective value, (b) droplet diameter, and (c) number of droplets per row for the Capillary number (Ca) and Weber number (We) conditions of each printed droplet pattern. Tuning the device's properties to achieve desired droplet conditions are illustrated here as a function of Ca and We.}

The parameter space envelope for each BO acquisition function shown in Figures \ref{fig:space} and \ref{fig:dimensionless}. For a microfluidics device, the slope of the linear ratio correlation will change depending on the properties of the device \cite{utada2007dripping, anna2006microscale}. The width of the instability drop formation region will change as a result of the fluids characteristics such as viscosity ratio. In the high-valued regions of Ca and We numbers, the oil pressure exceeds the viscosity ratio and, in turn, results in no water flow or water backflow. Moreover, in the low-valued regions of Ca and We numbers, the water pressure exceeds the viscosity ratio and forms a continuous water stream instead of droplets. In the specific case of ﬂow-focusing microsystem, the drop formation region can be divided into three distinguished sections. The first regime, "squeezing" or "geometric restriction", appears at the low flows/pressures and low capillary number area. The inner fluid is touching the restriction wall before breaking into individual drops \cite{garstecki2005mechanism}. In the squeezing regime, there is a good correlation between the drop size and the fluid flow fraction. The second regime, "dripping" or "transition" appearing at the mid-range flows/pressures and capillary number. In this regime, the formation of the droplets is due to an increased flow velocity inside the restrictions that leads to pinched off droplets. In this case, smaller droplets can be formed, and as we are increasing the fraction ratio between the inner and continuous liquids, a high volume of small drops is generated. The third regime is the "jetting" that appears at the highest pressures and capillary numbers, where the droplets form after the restriction.

\begin{figure}[p]
\centering
\begin{subfigure}{0.8\textwidth}
\includegraphics[width=\textwidth]{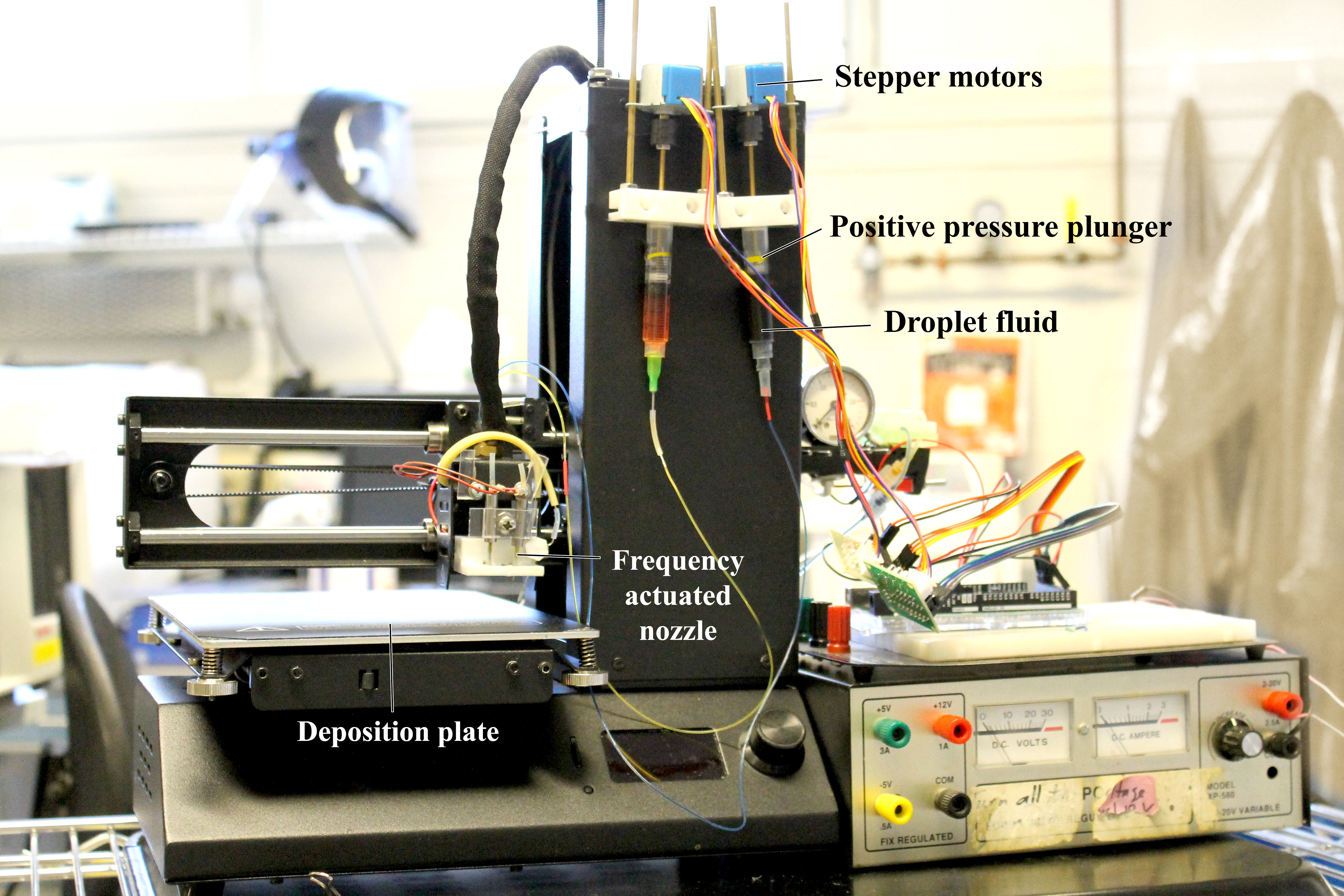}
\caption{\added{Inkjet Device Hardware}}
\end{subfigure}
\begin{subfigure}{0.55\textwidth}
\includegraphics[width=\textwidth]{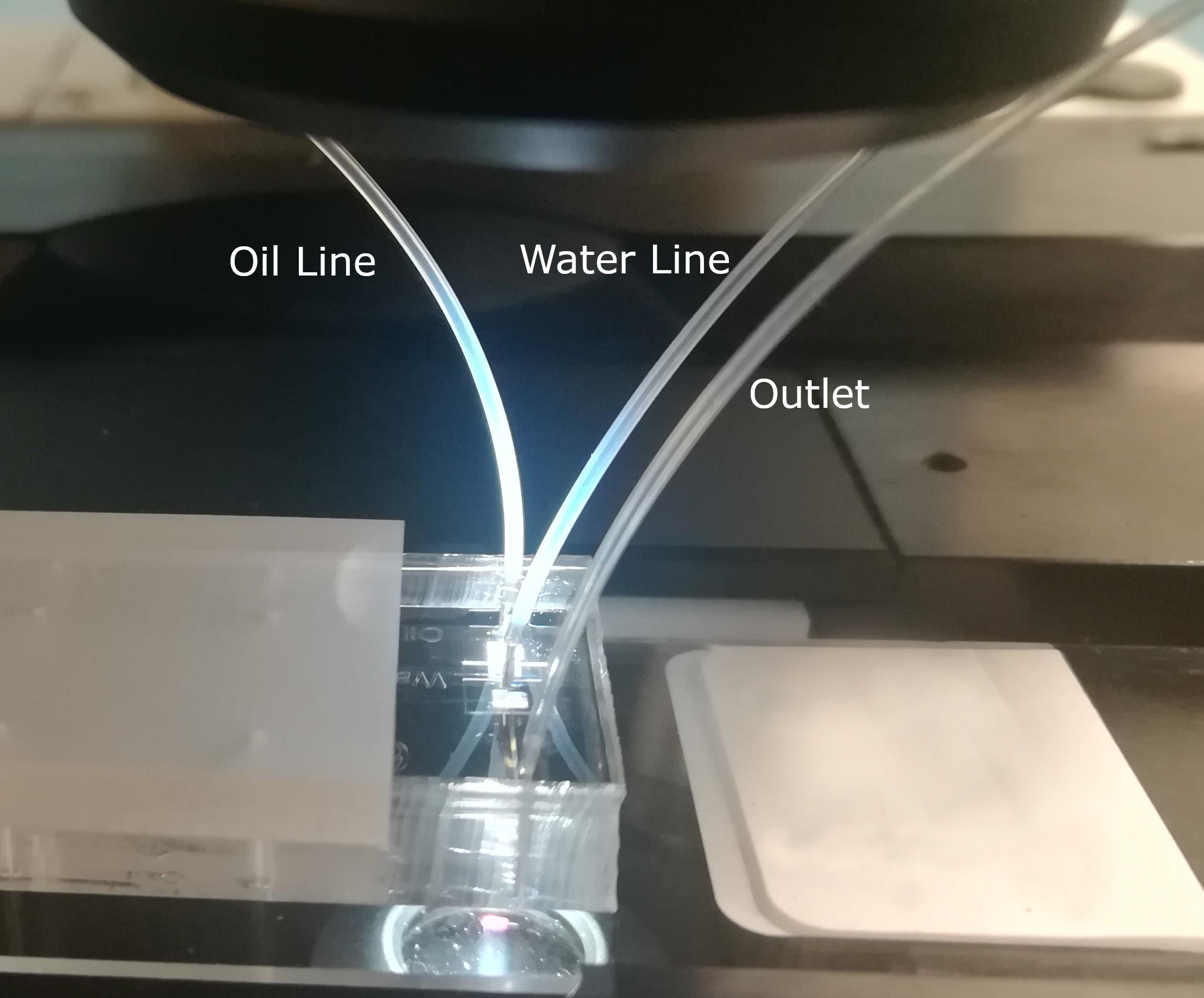}
\caption{\added{Microfluidics Device Hardware}}
\end{subfigure}
\caption{\added[id=AES, comment={Move the lab photos of device hardware to supplementary}]{Images of the (a) inkjet droplet-generator and the (b) microfluidics droplet-generator.}}
\label{fig:inkjet-diagram-lab}
\end{figure}

\footnotesize
\begin{longtable}[h]{cccccccc}
\label{table:inkjet_conds}\\
\caption{\added{Inkjet conditions for three independent Bayesian experiments of droplet structures using three different decision policies: EI, MPI, LCB. The initialization dataset is obtained via LHS. This table details all of the experiments conducted for the Bayesian experimentation and optimization of droplet structures using the inkjet droplet deposition system. The synthesis conditions of $N=3$ control parameters: pressure, frequency, and speed, all reported as normalized values. The last column is the computer vision-computed loss, $\ell$, for each measurement.}}\\
\toprule
 Sample &          Batch & Method &  Pressure &  Frequency &  Speed &  Loss \\
\midrule
      1 & Initialization &    LHS &     0.583 &      0.719 &  0.002 &      1.000 \\
      2 & Initialization &    LHS &     0.917 &      0.294 &  0.059 &      1.000 \\
      3 & Initialization &    LHS &     0.667 &      0.552 &  0.118 &      1.000 \\
      4 & Initialization &    LHS &     0.333 &      0.927 &  0.165 &      1.000 \\
      5 & Initialization &    LHS &     0.083 &      0.169 &  0.226 &      1.000 \\
      6 & Initialization &    LHS &     0.083 &      0.491 &  0.272 &      1.000 \\
      7 & Initialization &    LHS &     0.167 &      0.660 &  0.312 &      1.000 \\
      8 & Initialization &    LHS &     0.750 &      0.106 &  0.373 &      1.000 \\
      9 & Initialization &    LHS &     0.250 &      0.303 &  0.443 &      0.717 \\
     10 & Initialization &    LHS &     0.000 &      0.958 &  0.483 &      1.000 \\
     11 & Initialization &    LHS &     0.500 &      0.500 &  0.520 &      0.720 \\
     12 & Initialization &    LHS &     0.000 &      0.050 &  0.581 &      0.564 \\
     13 & Initialization &    LHS &     0.750 &      0.036 &  0.608 &      0.630 \\
     14 & Initialization &    LHS &     0.917 &      0.817 &  0.691 &      1.000 \\
     15 & Initialization &    LHS &     0.417 &      0.377 &  0.704 &      0.728 \\
     16 & Initialization &    LHS &     0.250 &      0.207 &  0.779 &      0.760 \\
     17 & Initialization &    LHS &     0.667 &      0.423 &  0.811 &      0.678 \\
     18 & Initialization &    LHS &     1.000 &      0.770 &  0.861 &      1.000 \\
     19 & Initialization &    LHS &     0.000 &      0.638 &  0.930 &      1.000 \\
     20 & Initialization &    LHS &     0.417 &      0.850 &  0.981 &      0.739 \\
     \midrule
     21 &        Batch 1 &     EI &     0.000 &      0.000 &  1.000 &      1.000 \\
     22 &        Batch 1 &     EI &     1.000 &      0.000 &  1.000 &      0.831 \\
     23 &        Batch 1 &     EI &     0.333 &      0.065 &  0.996 &      0.562 \\
     24 &        Batch 1 &     EI &     0.833 &      0.032 &  0.989 &      0.665 \\
     25 &        Batch 1 &     EI &     0.000 &      0.090 &  0.992 &      0.567 \\
     26 &        Batch 1 &     EI &     0.333 &      0.007 &  0.992 &      0.850 \\
     27 &        Batch 1 &     EI &     0.833 &      0.109 &  0.972 &      0.393 \\
     28 &        Batch 1 &     EI &     0.000 &      0.029 &  0.867 &      0.715 \\
     29 &        Batch 1 &     EI &     0.833 &      0.013 &  0.985 &      0.715 \\
     30 &        Batch 1 &     EI &     0.000 &      0.013 &  0.995 &      0.840 \\
     31 &        Batch 2 &     EI &     0.833 &      0.000 &  1.000 &      0.812 \\
     32 &        Batch 2 &     EI &     0.083 &      0.027 &  0.958 &      0.736 \\
     33 &        Batch 2 &     EI &     1.000 &      0.076 &  0.969 &      0.603 \\
     34 &        Batch 2 &     EI &     0.000 &      0.197 &  0.979 &      0.411 \\
     35 &        Batch 2 &     EI &     0.083 &      0.009 &  1.000 &      0.833 \\
     36 &        Batch 2 &     EI &     1.000 &      1.000 &  1.000 &      1.000 \\
     37 &        Batch 2 &     EI &     1.000 &      0.000 &  1.000 &      0.829 \\
     38 &        Batch 2 &     EI &     0.917 &      0.007 &  0.007 &      1.000 \\
     39 &        Batch 2 &     EI &     0.000 &      0.009 &  0.119 &      0.521 \\
     40 &        Batch 2 &     EI &     0.083 &      0.998 &  0.998 &      0.790 \\
     41 &        Batch 3 &     EI &     1.000 &      0.201 &  1.000 &      0.685 \\
     42 &        Batch 3 &     EI &     0.833 &      0.176 &  0.999 &      0.686 \\
     43 &        Batch 3 &     EI &     1.000 &      0.223 &  0.970 &      0.673 \\
     44 &        Batch 3 &     EI &     1.000 &      0.332 &  0.996 &      0.663 \\
     45 &        Batch 3 &     EI &     0.833 &      0.192 &  0.990 &      0.685 \\
     46 &        Batch 3 &     EI &     0.333 &      0.165 &  0.999 &      0.366 \\
     47 &        Batch 3 &     EI &     0.583 &      0.147 &  1.000 &      0.361 \\
     48 &        Batch 3 &     EI &     0.583 &      0.228 &  0.986 &      0.702 \\
     49 &        Batch 3 &     EI &     0.750 &      0.149 &  0.995 &      0.368 \\
     50 &        Batch 3 &     EI &     0.833 &      0.298 &  0.987 &      0.682 \\
     51 &        Batch 4 &     EI &     0.000 &      0.147 &  1.000 &      0.460 \\
     52 &        Batch 4 &     EI &     1.000 &      0.514 &  1.000 &      1.000 \\
     53 &        Batch 4 &     EI &     0.167 &      0.143 &  1.000 &      0.449 \\
     54 &        Batch 4 &     EI &     0.250 &      0.136 &  1.000 &      0.432 \\
     55 &        Batch 4 &     EI &     0.000 &      0.447 &  1.000 &      0.809 \\
     56 &        Batch 4 &     EI &     0.417 &      0.131 &  1.000 &      0.431 \\
     57 &        Batch 4 &     EI &     0.250 &      0.482 &  1.000 &      0.767 \\
     58 &        Batch 4 &     EI &     0.000 &      0.137 &  0.890 &      0.449 \\
     59 &        Batch 4 &     EI &     0.250 &      0.145 &  0.954 &      0.434 \\
     60 &        Batch 4 &     EI &     0.917 &      0.492 &  0.994 &      1.000 \\
     \midrule
     21 &        Batch 1 &    MPI &     0.417 &      0.000 &  1.000 &      0.885 \\
     22 &        Batch 1 &    MPI &     1.000 &      0.027 &  1.000 &      0.680 \\
     23 &        Batch 1 &    MPI &     0.083 &      0.036 &  0.907 &      0.704 \\
     24 &        Batch 1 &    MPI &     1.000 &      0.028 &  0.794 &      0.634 \\
     25 &        Batch 1 &    MPI &     0.000 &      0.066 &  0.986 &      0.640 \\
     26 &        Batch 1 &    MPI &     1.000 &      0.518 &  1.000 &      1.000 \\
     27 &        Batch 1 &    MPI &     0.000 &      0.058 &  0.523 &      0.516 \\
     28 &        Batch 1 &    MPI &     1.000 &      0.014 &  0.402 &      0.629 \\
     29 &        Batch 1 &    MPI &     0.000 &      0.476 &  0.999 &      0.809 \\
     30 &        Batch 1 &    MPI &     0.083 &      0.004 &  0.002 &      1.000 \\
     31 &        Batch 2 &    MPI &     0.000 &      0.000 &  0.644 &      0.877 \\
     32 &        Batch 2 &    MPI &     1.000 &      0.000 &  0.669 &      0.811 \\
     33 &        Batch 2 &    MPI &     0.000 &      0.000 &  0.879 &      0.915 \\
     34 &        Batch 2 &    MPI &     0.000 &      0.055 &  0.491 &      0.539 \\
     35 &        Batch 2 &    MPI &     0.000 &      0.001 &  0.748 &      0.915 \\
     36 &        Batch 2 &    MPI &     0.750 &      0.060 &  0.826 &      0.510 \\
     37 &        Batch 2 &    MPI &     0.500 &      1.000 &  0.717 &      1.000 \\
     38 &        Batch 2 &    MPI &     0.000 &      0.048 &  0.597 &      0.474 \\
     39 &        Batch 2 &    MPI &     0.917 &      0.027 &  0.715 &      0.617 \\
     40 &        Batch 2 &    MPI &     0.000 &      0.009 &  0.514 &      0.766 \\
     41 &        Batch 3 &    MPI &     0.917 &      0.000 &  0.658 &      0.811 \\
     42 &        Batch 3 &    MPI &     0.000 &      0.037 &  0.556 &      0.689 \\
     43 &        Batch 3 &    MPI &     0.417 &      0.010 &  0.660 &      0.686 \\
     44 &        Batch 3 &    MPI &     1.000 &      0.022 &  0.948 &      0.681 \\
     45 &        Batch 3 &    MPI &     0.000 &      0.029 &  0.799 &      1.000 \\
     46 &        Batch 3 &    MPI &     1.000 &      0.141 &  0.414 &      1.000 \\
     47 &        Batch 3 &    MPI &     0.000 &      0.672 &  0.584 &      1.000 \\
     48 &        Batch 3 &    MPI &     1.000 &      1.000 &  1.000 &      1.000 \\
     49 &        Batch 3 &    MPI &     0.083 &      0.016 &  0.304 &      0.605 \\
     50 &        Batch 3 &    MPI &     1.000 &      0.079 &  0.981 &      0.511 \\
     51 &        Batch 4 &    MPI &     0.833 &      0.000 &  1.000 &      0.885 \\
     52 &        Batch 4 &    MPI &     0.083 &      0.019 &  0.905 &      0.759 \\
     53 &        Batch 4 &    MPI &     0.917 &      0.001 &  0.013 &      1.000 \\
     54 &        Batch 4 &    MPI &     0.000 &      0.086 &  0.962 &      0.636 \\
     55 &        Batch 4 &    MPI &     1.000 &      0.097 &  0.996 &      0.430 \\
     56 &        Batch 4 &    MPI &     0.083 &      0.011 &  0.029 &      1.000 \\
     57 &        Batch 4 &    MPI &     1.000 &      0.005 &  0.226 &      0.622 \\
     58 &        Batch 4 &    MPI &     0.083 &      0.053 &  0.948 &      1.000 \\
     59 &        Batch 4 &    MPI &     1.000 &      1.000 &  1.000 &      1.000 \\
     60 &        Batch 4 &    MPI &     0.083 &      0.003 &  0.045 &      1.000 \\
    \midrule
     21 &        Batch 1 &    LCB &     0.667 &      0.000 &  1.000 &      0.739 \\
     22 &        Batch 1 &    LCB &     0.500 &      0.077 &  0.996 &      0.529 \\
     23 &        Batch 1 &    LCB &     0.833 &      0.322 &  0.998 &      0.665 \\
     24 &        Batch 1 &    LCB &     0.917 &      0.053 &  0.980 &      0.610 \\
     25 &        Batch 1 &    LCB &     0.250 &      0.038 &  0.944 &      0.692 \\
     26 &        Batch 1 &    LCB &     0.500 &      0.015 &  0.550 &      0.655 \\
     27 &        Batch 1 &    LCB &     0.500 &      0.422 &  0.996 &      1.000 \\
     28 &        Batch 1 &    LCB &     0.083 &      0.021 &  0.554 &      0.692 \\
     29 &        Batch 1 &    LCB &     0.250 &      0.020 &  0.520 &      0.656 \\
     30 &        Batch 1 &    LCB &     0.750 &      0.702 &  1.000 &      1.000 \\
     31 &        Batch 2 &    LCB &     0.000 &      0.000 &  1.000 &      1.000 \\
     32 &        Batch 2 &    LCB &     0.917 &      0.037 &  0.118 &      1.000 \\
     33 &        Batch 2 &    LCB &     1.000 &      1.000 &  1.000 &      1.000 \\
     34 &        Batch 2 &    LCB &     0.000 &      0.910 &  0.021 &      1.000 \\
     35 &        Batch 2 &    LCB &     0.833 &      0.054 &  0.985 &      0.569 \\
     36 &        Batch 2 &    LCB &     0.000 &      0.021 &  0.177 &      0.470 \\
     37 &        Batch 2 &    LCB &     0.000 &      0.969 &  0.907 &      1.000 \\
     38 &        Batch 2 &    LCB &     1.000 &      0.996 &  0.107 &      1.000 \\
     39 &        Batch 2 &    LCB &     1.000 &      0.023 &  0.831 &      0.699 \\
     40 &        Batch 2 &    LCB &     0.083 &      0.009 &  0.963 &      0.815 \\
     41 &        Batch 3 &    LCB &     0.000 &      0.000 &  1.000 &      1.000 \\
     42 &        Batch 3 &    LCB &     0.917 &      0.030 &  0.006 &      1.000 \\
     43 &        Batch 3 &    LCB &     1.000 &      1.000 &  1.000 &      1.000 \\
     44 &        Batch 3 &    LCB &     0.000 &      0.735 &  0.003 &      1.000 \\
     45 &        Batch 3 &    LCB &     1.000 &      0.032 &  0.914 &      0.607 \\
     46 &        Batch 3 &    LCB &     0.083 &      0.997 &  0.978 &      1.000 \\
     47 &        Batch 3 &    LCB &     0.083 &      0.027 &  0.021 &      1.000 \\
     48 &        Batch 3 &    LCB &     0.917 &      0.947 &  0.038 &      1.000 \\
     49 &        Batch 3 &    LCB &     0.083 &      0.006 &  0.962 &      0.666 \\
     50 &        Batch 3 &    LCB &     1.000 &      0.160 &  0.004 &      1.000 \\
     51 &        Batch 4 &    LCB &     1.000 &      0.000 &  1.000 &      0.827 \\
     52 &        Batch 4 &    LCB &     0.083 &      0.071 &  0.542 &      0.464 \\
     53 &        Batch 4 &    LCB &     1.000 &      1.000 &  0.000 &      1.000 \\
     54 &        Batch 4 &    LCB &     0.083 &      0.930 &  0.964 &      1.000 \\
     55 &        Batch 4 &    LCB &     0.833 &      0.006 &  0.060 &      1.000 \\
     56 &        Batch 4 &    LCB &     1.000 &      1.000 &  1.000 &      1.000 \\
     57 &        Batch 4 &    LCB &     0.000 &      0.869 &  0.068 &      1.000 \\
     58 &        Batch 4 &    LCB &     0.250 &      0.065 &  0.992 &      0.569 \\
     59 &        Batch 4 &    LCB &     0.000 &      0.050 &  0.016 &      1.000 \\
     60 &        Batch 4 &    LCB &     0.917 &      0.009 &  0.848 &      0.746 \\
\bottomrule
\end{longtable}

\begin{longtable}[h]{cccccc}
\caption{\added{Microfluidics conditions for three independent Bayesian experiments of droplet structures using three different decision policies: EI, MPI, LCB. The initialization dataset is obtained via LHS. This table details all of the experiments conducted for the Bayesian experimentation and optimization of droplet structures using the microfluidic droplet deposition system. The synthesis conditions of $N=2$ control parameters: water pressure and oil pressure, all displayed as normalized values. The last column is the computer vision-computed loss, $\ell$, for each measurement.}}\\
\toprule
 Sample &          Batch & Method &  Water Pressure &  Oil Pressure &  Loss \\
\midrule
      1 & Initialization &    LHS &          0.420 &        0.040 &      1.000 \\
      2 & Initialization &    LHS &          0.043 &        0.083 &      0.352 \\
      3 & Initialization &    LHS &          0.255 &        0.138 &      1.000 \\
      4 & Initialization &    LHS &          0.727 &        0.168 &      1.000 \\
      5 & Initialization &    LHS &          0.870 &        0.218 &      1.000 \\
      6 & Initialization &    LHS &          0.155 &        0.277 &      0.365 \\
      7 & Initialization &    LHS &          0.141 &        0.333 &      0.319 \\
      8 & Initialization &    LHS &          0.812 &        0.399 &      1.000 \\
      9 & Initialization &    LHS &          0.760 &        0.442 &      1.000 \\
     10 & Initialization &    LHS &          0.948 &        0.465 &      1.000 \\
     11 & Initialization &    LHS &          0.303 &        0.549 &      0.320 \\
     12 & Initialization &    LHS &          0.208 &        0.599 &      0.291 \\
     13 & Initialization &    LHS &          0.988 &        0.618 &      1.000 \\
     14 & Initialization &    LHS &          0.575 &        0.674 &      1.000 \\
     15 & Initialization &    LHS &          0.540 &        0.713 &      1.000 \\
     16 & Initialization &    LHS &          0.648 &        0.796 &      1.000 \\
     17 & Initialization &    LHS &          0.478 &        0.824 &      1.000 \\
     18 & Initialization &    LHS &          0.068 &        0.887 &      1.000 \\
     19 & Initialization &    LHS &          0.683 &        0.916 &      1.000 \\
     20 & Initialization &    LHS &          0.392 &        0.979 &      0.196 \\
     \midrule
     21 &        Batch 1 &     EI &          0.305 &        1.000 &      0.145 \\
     22 &        Batch 1 &     EI &          0.321 &        0.987 &      0.157 \\
     23 &        Batch 1 &     EI &          0.297 &        0.993 &      0.080 \\
     24 &        Batch 1 &     EI &          0.315 &        0.949 &      0.176 \\
     25 &        Batch 1 &     EI &          0.290 &        0.964 &      0.167 \\
     26 &        Batch 1 &     EI &          0.302 &        0.915 &      0.188 \\
     27 &        Batch 1 &     EI &          0.261 &        0.999 &      0.073 \\
     28 &        Batch 1 &     EI &          0.330 &        0.992 &      0.186 \\
     29 &        Batch 1 &     EI &          0.285 &        0.936 &      0.172 \\
     30 &        Batch 1 &     EI &          0.294 &        0.881 &      0.194 \\
     31 &        Batch 2 &     EI &          0.228 &        1.000 &      0.132 \\
     32 &        Batch 2 &     EI &          0.219 &        0.989 &      0.137 \\
     33 &        Batch 2 &     EI &          0.214 &        0.924 &      0.134 \\
     34 &        Batch 2 &     EI &          0.203 &        0.997 &      0.201 \\
     35 &        Batch 2 &     EI &          0.227 &        0.948 &      0.168 \\
     36 &        Batch 2 &     EI &          1.000 &        0.000 &      1.000 \\
     37 &        Batch 2 &     EI &          0.176 &        0.995 &      0.242 \\
     38 &        Batch 2 &     EI &          0.243 &        0.987 &      0.170 \\
     39 &        Batch 2 &     EI &          0.202 &        0.899 &      0.168 \\
     40 &        Batch 2 &     EI &          0.260 &        0.995 &      0.131 \\
     41 &        Batch 3 &     EI &          0.261 &        1.000 &      0.090 \\
     42 &        Batch 3 &     EI &          0.260 &        0.977 &      0.115 \\
     43 &        Batch 3 &     EI &          0.272 &        0.999 &      0.105 \\
     44 &        Batch 3 &     EI &          0.099 &        0.000 &      1.000 \\
     45 &        Batch 3 &     EI &          0.002 &        0.000 &      1.000 \\
     46 &        Batch 3 &     EI &          0.251 &        0.992 &      0.111 \\
     47 &        Batch 3 &     EI &          0.879 &        1.000 &      1.000 \\
     48 &        Batch 3 &     EI &          0.254 &        0.987 &      0.097 \\
     49 &        Batch 3 &     EI &          0.258 &        0.935 &      0.153 \\
     50 &        Batch 3 &     EI &          0.270 &        0.960 &      0.151 \\
     51 &        Batch 4 &     EI &          0.000 &        0.252 &      1.000 \\
     52 &        Batch 4 &     EI &          0.009 &        0.234 &      1.000 \\
     53 &        Batch 4 &     EI &          0.019 &        0.316 &      1.000 \\
     54 &        Batch 4 &     EI &          0.031 &        0.188 &      0.971 \\
     55 &        Batch 4 &     EI &          0.005 &        0.372 &      1.000 \\
     56 &        Batch 4 &     EI &          0.003 &        0.296 &      1.000 \\
     57 &        Batch 4 &     EI &          0.030 &        0.199 &      1.000 \\
     58 &        Batch 4 &     EI &          0.001 &        0.197 &      1.000 \\
     59 &        Batch 4 &     EI &          0.049 &        0.251 &      0.483 \\
     60 &        Batch 4 &     EI &          0.003 &        0.166 &      1.000 \\
     \midrule
     21 &        Batch 1 &    MPI &          0.382 &        0.989 &      0.180 \\
     22 &        Batch 1 &    MPI &          0.360 &        0.947 &      0.194 \\
     23 &        Batch 1 &    MPI &          0.327 &        0.886 &      0.194 \\
     24 &        Batch 1 &    MPI &          0.332 &        0.996 &      0.102 \\
     25 &        Batch 1 &    MPI &          0.316 &        0.818 &      0.216 \\
     26 &        Batch 1 &    MPI &          0.316 &        0.938 &      0.163 \\
     27 &        Batch 1 &    MPI &          0.316 &        0.744 &      0.261 \\
     28 &        Batch 1 &    MPI &          0.281 &        0.861 &      0.178 \\
     29 &        Batch 1 &    MPI &          0.371 &        0.922 &      0.203 \\
     30 &        Batch 1 &    MPI &          0.295 &        0.995 &      0.132 \\
     31 &        Batch 2 &    MPI &          0.335 &        1.000 &      0.188 \\
     32 &        Batch 2 &    MPI &          0.334 &        0.979 &      0.180 \\
     33 &        Batch 2 &    MPI &          0.291 &        0.997 &      0.079 \\
     34 &        Batch 2 &    MPI &          0.313 &        0.994 &      0.180 \\
     35 &        Batch 2 &    MPI &          0.266 &        0.982 &      0.089 \\
     36 &        Batch 2 &    MPI &          0.324 &        0.999 &      0.191 \\
     37 &        Batch 2 &    MPI &          0.325 &        0.998 &      0.125 \\
     38 &        Batch 2 &    MPI &          0.357 &        0.987 &      0.202 \\
     39 &        Batch 2 &    MPI &          0.356 &        0.993 &      0.191 \\
     40 &        Batch 2 &    MPI &          0.228 &        0.998 &      0.164 \\
     41 &        Batch 3 &    MPI &          0.268 &        1.000 &      0.095 \\
     42 &        Batch 3 &    MPI &          0.246 &        0.962 &      0.091 \\
     43 &        Batch 3 &    MPI &          0.281 &        0.999 &      0.088 \\
     44 &        Batch 3 &    MPI &          0.270 &        0.953 &      0.109 \\
     45 &        Batch 3 &    MPI &          0.279 &        0.967 &      0.118 \\
     46 &        Batch 3 &    MPI &          0.260 &        0.921 &      0.166 \\
     47 &        Batch 3 &    MPI &          0.261 &        0.989 &      0.103 \\
     48 &        Batch 3 &    MPI &          0.289 &        0.992 &      0.123 \\
     49 &        Batch 3 &    MPI &          0.260 &        0.969 &      0.110 \\
     50 &        Batch 3 &    MPI &          0.253 &        0.890 &      0.135 \\
     51 &        Batch 4 &    MPI &          0.265 &        1.000 &      0.107 \\
     52 &        Batch 4 &    MPI &          0.276 &        1.000 &      0.070 \\
     53 &        Batch 4 &    MPI &          0.276 &        0.989 &      0.047 \\
     54 &        Batch 4 &    MPI &          0.274 &        0.974 &      0.082 \\
     55 &        Batch 4 &    MPI &          0.262 &        1.000 &      0.097 \\
     56 &        Batch 4 &    MPI &          0.253 &        0.957 &      0.139 \\
     57 &        Batch 4 &    MPI &          0.000 &        0.000 &      1.000 \\
     58 &        Batch 4 &    MPI &          0.252 &        0.980 &      0.124 \\
     59 &        Batch 4 &    MPI &          0.257 &        0.986 &      0.069 \\
     60 &        Batch 4 &    MPI &          0.268 &        0.952 &      0.141 \\
    \midrule
     21 &        Batch 1 &    LCB &          1.000 &        0.000 &      1.000 \\
     22 &        Batch 1 &    LCB &          0.000 &        0.503 &      1.000 \\
     23 &        Batch 1 &    LCB &          0.000 &        0.280 &      1.000 \\
     24 &        Batch 1 &    LCB &          0.000 &        0.000 &      1.000 \\
     25 &        Batch 1 &    LCB &          0.254 &        1.000 &      0.097 \\
     26 &        Batch 1 &    LCB &          0.001 &        0.172 &      1.000 \\
     27 &        Batch 1 &    LCB &          1.000 &        1.000 &      1.000 \\
     28 &        Batch 1 &    LCB &          0.142 &        0.495 &      0.316 \\
     29 &        Batch 1 &    LCB &          0.006 &        0.639 &      1.000 \\
     30 &        Batch 1 &    LCB &          0.452 &        1.000 &      0.250 \\
     31 &        Batch 2 &    LCB &          0.301 &        1.000 &      0.083 \\
     32 &        Batch 2 &    LCB &          0.277 &        0.846 &      0.107 \\
     33 &        Batch 2 &    LCB &          0.226 &        0.661 &      0.276 \\
     34 &        Batch 2 &    LCB &          0.161 &        0.997 &      0.430 \\
     35 &        Batch 2 &    LCB &          0.000 &        0.125 &      1.000 \\
     36 &        Batch 2 &    LCB &          0.178 &        0.428 &      0.319 \\
     37 &        Batch 2 &    LCB &          0.373 &        0.950 &      0.207 \\
     38 &        Batch 2 &    LCB &          0.149 &        0.204 &      0.443 \\
     39 &        Batch 2 &    LCB &          0.367 &        0.761 &      0.304 \\
     40 &        Batch 2 &    LCB &          0.335 &        0.534 &      1.000 \\
     41 &        Batch 3 &    LCB &          0.298 &        1.000 &      0.150 \\
     42 &        Batch 3 &    LCB &          0.246 &        0.875 &      0.094 \\
     43 &        Batch 3 &    LCB &          0.225 &        0.724 &      0.253 \\
     44 &        Batch 3 &    LCB &          0.394 &        0.991 &      0.192 \\
     45 &        Batch 3 &    LCB &          0.180 &        0.522 &      0.303 \\
     46 &        Batch 3 &    LCB &          0.195 &        0.993 &      0.283 \\
     47 &        Batch 3 &    LCB &          0.133 &        0.294 &      0.333 \\
     48 &        Batch 3 &    LCB &          0.121 &        0.000 &      1.000 \\
     49 &        Batch 3 &    LCB &          0.358 &        0.841 &      0.171 \\
     50 &        Batch 3 &    LCB &          0.303 &        0.607 &      0.331 \\
     51 &        Batch 4 &    LCB &          0.307 &        1.000 &      0.133 \\
     52 &        Batch 4 &    LCB &          0.264 &        0.812 &      0.236 \\
     53 &        Batch 4 &    LCB &          0.192 &        0.619 &      0.285 \\
     54 &        Batch 4 &    LCB &          0.216 &        0.993 &      0.189 \\
     55 &        Batch 4 &    LCB &          0.165 &        0.422 &      0.323 \\
     56 &        Batch 4 &    LCB &          0.431 &        0.999 &      0.310 \\
     57 &        Batch 4 &    LCB &          0.321 &        0.695 &      0.306 \\
     58 &        Batch 4 &    LCB &          0.141 &        0.250 &      0.373 \\
     59 &        Batch 4 &    LCB &          0.332 &        0.895 &      0.212 \\
     60 &        Batch 4 &    LCB &          0.141 &        0.807 &      0.477 \\
\bottomrule
\end{longtable}

\end{document}